\pdfoutput=1

\documentclass[11pt]{article}

\usepackage[]{acl}

\usepackage[T1]{fontenc}

\usepackage[utf8]{inputenc}

\usepackage{microtype}
\usepackage{times}
\usepackage{latexsym}
\usepackage{amssymb}
\usepackage{amsfonts}
\usepackage{amsmath} 
\usepackage{amsthm} 
\usepackage{booktabs}
\usepackage{enumerate}
\usepackage{enumitem}
\usepackage{graphicx}
\usepackage{subfigure}
\usepackage{xspace}
\usepackage{float}
\usepackage{bbm}
\usepackage{bm}
\usepackage{multirow}
\usepackage{booktabs}
\usepackage{color}
\usepackage{framed}
\usepackage{stfloats}
\usepackage{iitem}
\usepackage{makecell}
\usepackage{float}
\usepackage{placeins}
\usepackage{afterpage}
\usepackage{bbding}
\usepackage[normalem]{ulem}
\usepackage{soul}
\usepackage{xcolor}
\usepackage{color}
\usepackage{algorithm}
\usepackage{algorithmic}
\usepackage{array}
\usepackage{hyperref}

\newcommand{\ctext}[3][RGB]{%
  \begingroup
  \definecolor{hlcolor}{#1}{#2}\sethlcolor{hlcolor}%
  \hl{#3}%
  \endgroup
}

\usepackage{url}
\newcommand{\paratitle}[1]{\vspace{1.5ex}\noindent\textbf{#1}}
\newcommand{\ie}{\emph{i.e.,}\xspace}

\newcommand{\eg}{\emph{e.g.,}\xspace}

\newcommand{\ignore}[1]{}
\newcommand{\tabincell}[2]{\begin{tabular}{@{}#1@{}}#2\end{tabular}}

\usepackage{xcolor}
\definecolor{tOrange}{RGB}{255,165,0}
\definecolor{tBlue}{rgb}{0.39,0.58,0.93}
\definecolor{tPink}{RGB}{255,20,147}
\definecolor{tGreen}{RGB}{50,205,50}
\definecolor{tGold}{RGB}{255,215,0}
\newcolumntype{P}[1]{>{\centering\arraybackslash}p{#1}}
\newcolumntype{M}[1]{>{\centering\arraybackslash}m{#1}}

\title{The Web Can Be Your Oyster for Improving Large Language Models}

\author{
	Junyi Li\textsuperscript{\rm{1,3,5}}, 
	Tianyi Tang\textsuperscript{\rm{1}},
        Wayne Xin Zhao\textsuperscript{\rm{1,5}\thanks{\ \ Corresponding author}\ }, 
        Jingyuan Wang\textsuperscript{\rm{4}\ }, \\ 
	\textbf{Jian-Yun Nie}\textsuperscript{\rm{3}} {\rm and} 
	\textbf{Ji-Rong Wen}\textsuperscript{\rm{1,2,5}} \\
	\textsuperscript{1}Gaoling School of Artificial Intelligence, Renmin University of China \\
	\textsuperscript{2}School of Information, Renmin University of China \\
	\textsuperscript{3}DIRO, Universit\'{e} de Montr\'{e}al \qquad \textsuperscript{4}Beihang University \\
	\textsuperscript{5}Beijing Key Laboratory of Big Data Management and Analysis Methods \\
	\texttt{lijunyi@ruc.edu.cn} \quad
        \texttt{steventianyitang@outlook.com} \quad
	\texttt{batmanfly@gmail.com} \\
}

\begin{document}
\maketitle

\begin{abstract}
Large language models (LLMs) encode a large amount of world knowledge. However, as such knowledge is frozen at the time of model training, the models become  \emph{static} and \emph{limited} by the training data at that time. 
In order to further improve the  capacity of LLMs for knowledge-intensive tasks, we consider augmenting LLMs with the large-scale web using search engine. Unlike previous augmentation sources (\eg Wikipedia data dump), the web provides  broader, more comprehensive and constantly updated information. 
In this paper, we present a web-augmented LLM -- \textsc{UniWeb}, which is trained over 16 knowledge-intensive tasks in a unified text-to-text format. Instead of simply using the retrieved contents from web, our approach has made two major improvements. Firstly, we propose an adaptive \emph{search engine assisted learning}  method that can self-evaluate the confidence level of LLM’s predictions, and adaptively determine when to refer to the web for more data, which can avoid useless or noisy augmentation from web. Secondly, we  design a pretraining task, \ie \emph{continual knowledge learning}, based on salient spans prediction, to reduce the discrepancy between the encoded and retrieved knowledge. Experiments on a wide range of knowledge-intensive tasks show that our model significantly outperforms previous retrieval-augmented methods. Our code and data can be accessed at this link \url{https://github.com/RUCAIBox/UniWeb}
\end{abstract}

\section{Introduction}
\label{sec-intro}

With large-scale neural networks, large language models~(LLMs)~\cite{gpt3,LLM-survey} can encode a large amount of world knowledge, showing  
phenomenal capability
in knowledge-intensive tasks such as fact checking
and open-domain question answering~(QA). However, this capacity is  naturally limited by the information contained in pretraining or finetuning datasets (usually fixed once collected), which are neither \emph{up-to-date} nor \emph{complete}~\cite{komeili2021internet,ji2022survey}. Although model scaling~\cite{gpt3,chowdhery2022palm,thoppilan2022lamda} is a viable way to improve the knowledge capacity of LLMs, it still uses \emph{static} pretraining datasets, and also leads to significantly larger  computational costs with increased model sizes. As a result, the outdated or incomplete
knowledge encoded by LLMs may lead to hallucination or incorrect generations  even though the results look plausible~\cite{ji2022survey}.

\begin{table}[t]
\renewcommand\arraystretch{1.1}
\setlength\tabcolsep{3pt}
	\centering
	\small
	\begin{tabular}{r p{0.35\textwidth}}
		\toprule[1pt]
		\textbf{Question} & Which popular \textbf{Korean show} was \textbf{recently} green lit for a new season? \\
            \textbf{Answer} & Squid Game \\
		\midrule[0.5pt]
		\textbf{Wikipedia} & \textcolor{red}{There are no results for the question.} \\
            \midrule[0.5pt]
            \textbf{Web} & 
            [...] Netflix announce Sunday that \ctext[RGB]{97,189,252}{the wildly} \\ 
            \multicolumn{2}{p{0.465\textwidth}}{
            \ctext[RGB]{97,189,252}{popular South Korean show is green lit for a second season.} ``And now, Gi-hun returns'' ``The Front Man returns. \ctext[RGB]{97,189,252}{Season 2 is coming.'' ``Squid Game'' is a fictional drama from South Korea} in which contestants who are desperately in need of money play deadly children's games to win cash prizes.  [...]} \\
            URL: & \url{https://www.cnn.com/2022/06/1} \\
            \multicolumn{2}{p{0.465\textwidth}}{\url{2/media/squid-game-season-2/index.html}} \\
            \midrule[0.5pt]
            \textbf{T5 w/o Web} & The Walking Dead \textcolor{red}{\XSolidBrush} \\
            \textbf{T5 w/ Web} & Squid Game \textcolor{blue}{\CheckmarkBold} \\
		\bottomrule[1pt]
	\end{tabular}%
	\caption{An example showing that the web covers both more comprehensive (\eg Korean show) and up-to-date (\eg recently) information than Wikipedia. Based on the latest news returned by Google Search, T5-\textsc{Large} can answer the question correctly.}
        \label{tab-example}%
        \vspace{-0.3cm}
\end{table}

Recently, by drawing the idea from semi-parametric approaches~\cite{dense-survey,realm,rag,borgeaud2022improving}, retrieval-augmented approaches have been proposed to equip LLMs with the ability to directly access an external database. As a major knowledge resource,  Wikipedia has been widely used in previous work. While being highly accurate and well-structured, Wikipedia
only covers \emph{limited} information, both in scope and in time. Besides, even for the topics that Wikipedia covers, grounding LLMs' decisions on a single source of knowledge may create biases~\cite{wagner2016women}.
Considering these issues, it is time to look beyond Wikipedia (or similar single-source databases) and access more \emph{broader}, \emph{in-depth}, and \emph{up-to-date} knowledge from more sources. 
Inspired by \cite{komeili2021internet,sphere}, we select the \emph{web} as the retrieval resource for enlarging the knowledge capacity of LLMs. 
To motivate our approach, Table~\ref{tab-example} presents a sample question that T5 successfully answers with the support of the web (providing the latest news), but not Wikipedia.
As we can see, timely and relevant supporting evidence is the key to solve such tasks for LLMs.   


In this paper, we aim to capitalize on the web as a source of up-to-date and comprehensive knowledge  to solve a wide range of knowledge-intensive tasks.
Unlike previous web-augmented studies~\cite{webgpt,menick2022teaching} that mostly focus on single tasks, we seek to develop a unified framework to integrate the use of the web in LLMs for multi-task solving.  
Although the idea of leveraging the web for improving LLMs is appealing, it is non-trivial to develop  an effective solution. First, LLMs do not always need external evidence for task solving, especially considering the fact that the web contains noisy,  biased, or harmful information~\cite{luccioni2021s}. Simply retrieving knowledge without considering the example difficulty and LLMs' own capabilities may steer models towards unexpected outputs. Second, LLMs are usually pretrained at an earlier time on a limited corpus, leading to a discrepancy between the encoded knowledge and the retrieved knowledge (\ie web contents). Therefore, we need more principled approaches to properly integrating the new knowledge into LLMs. 


To address the above issues, we present 
a web-augmented LLMs, \textsc{UniWeb}, to improve the capacity in  knowledge-intensive tasks. Instead of using neural network-based  retriever, we employ a commercial  search engine (\ie Google Search) to obtain high-quality and comprehensive retrieval results from the web. 
Based on this idea, 
we make two major technical contributions. 
First, we propose a \textit{search engine assisted learning} method that can selectively query the web only when LLM is unconfident in its predictions.  For this purpose, we design a self-evaluation mechanism to estimate the confidence level of LLMs on  the task examples.  Secondly, to reduce the discrepancy between the encoded and retrieved knowledge,  we design a pre-training task, \textit{continual knowledge learning}, to integrate the retrieved knowledge into LLMs by predicting the salient masked spans in web documents.
To train the \textsc{UniWeb} model, we convert different knowledge-intensive tasks into a unified text-to-text format, and conduct  supervised multi-task training over 16 tasks across seven categories. 


To the best of our knowledge, our model is the first unified web-augmented LLM for a wide range of knowledge-intensive tasks. Extensive experiments show that  LLMs can significantly benefit from such an approach and a single unified LLM (\textsc{UniWeb}) is able to achieve (near) state-of-the-art performance on all 16 tasks. 

\section{Related Work}

\paratitle{Retrieval-Augmented LLMs.} Augmenting a large language model with retrieval has been extensively studied in existing literature~\cite{rag,borgeaud2022improving,izacard2022few,orqa,realm}. For example, REALM~\cite{realm} and RAG~\cite{rag}, incorporate a differentiable retriever into pretrained models, leading to promising results on question answering. However, these studies usually rely on a sub-optimal retriever to access a static and limited knowledge resource, \ie Wikipedia. By contrast, our model utilizes the well-developed search engine to gain broader, more in-depth, and up-to-date knowledge from the web. Several studies have also looked at how Internet can help the models, but only focus on single tasks such as question answering~\cite{webgpt,menick2022teaching} and dialogue~\cite{komeili2021internet}.
WebGPT~\cite{webgpt} uses human feedback to optimize answer quality by hiring massive labelers to judge the accuracy of answers. \citet{komeili2021internet} retrieves knowledge from the web for every dialogue without considering the necessity. \citet{sphere} only presents an empirical study to investigate the impact of replacing Wikipedia with a large-scale web-like corpus and adopting different retrieval models. We are also aware of some related studies~\cite{jiang2023active}, but we have taken a different active approach for knowledge retrieval. In this paper, we develop a unified language model for solving a wide spectrum of knowledge-intensive tasks. Our model can selectively decide whether to access the web, and continuously learn from the retrieved knowledge.


\paratitle{Knowledge-Intensive Learning.} Recent work has shown that LLMs' parameters have implicitly stored linguistic or factual knowledge~\cite{petroni2019language,roberts2020much}. However, the implicitly encoded knowledge is limited by the model's scale and training data, contradicting the dynamic nature of the world. Hence, many researchers propose to fuse relevant external knowledge from texts with the encoded knowledge of LLMs to deal with knowledge-intensive tasks such as open-domain QA~\cite{realm,rag}, entity linking~\cite{wu2019scalable}, fact verification~\cite{liu2019fine}, and commonsense
reasoning~\cite{numersense}. Wikipedia has been the most widely used knowledge source for these tasks, which is still limited despite its wide coverage. Instead, we rely on the real-time web. The existing studies usually design task-specific training, architecture, and knowledge fusion method to exploit knowledge sources. 
In this work, we aim to develop a single unified framework that can be used for most knowledge-intensive tasks.

\section{Task Formulation}
\label{sec-pre}

\begin{figure*}[tb]
	\centering
	\includegraphics[width=1.0\textwidth]{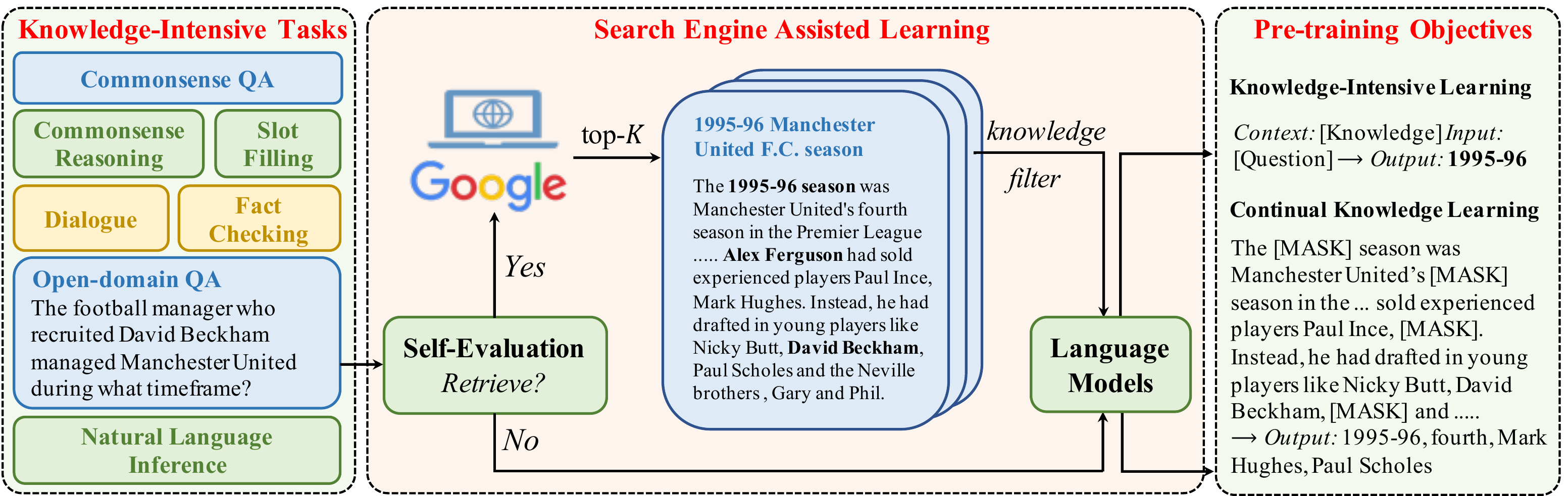}
	\caption{Overview of our proposed web-augmented large language model \textsc{UniWeb}.}
	\label{fig-model}
	\vspace{-0.2cm}
\end{figure*}

Knowledge-intensive tasks~\cite{kit_survey} aim to leverage external knowledge resources to accomplish a broad range of tasks such as open-domain question answering and fact verification.

Following prior  work~\cite{rag,realm}, we employ a retrieval-augmented generation framework that consists of two components: a retriever $\mathcal{R}$ and a generator $\mathcal{G}$. Given an input text $\mathcal{X}$ such as a question, the retriever $\mathcal{R}$ learns to retrieve a set of top-$K$ passages $\mathcal{P}=\{p_1,...,p_K\}$ from a knowledge resource. Conditioned on the input text $\mathcal{X}$ and the retrieved passages $\mathcal{P}$, the generator $\mathcal{G}$ aims to generate the output text $\mathcal{Y}$. The model is trained to maximize the joint likelihood: 
\begin{equation}
    \text{Pr}(\mathcal{Y}|\mathcal{X}) = \sum_{\mathcal{R},\mathcal{G}} \text{Pr}(\mathcal{P}|\mathcal{X}) \text{Pr}(\mathcal{Y}|\mathcal{P},\mathcal{X}).
\end{equation}

To implement the framework, 
previous studies usually adopt a trainable neural retriever based on  a (single) knowledge resource such as Wikipedia or knowledge bases. However, such an approach can only access limited, static knowledge. 
In this paper, we rely on a general, off-the-shelf search engine  as the retriever to access both \textit{comprehensive} and \textit{up-to-date} knowledge from the whole web.

\section{Approach}

Our proposed web-augmented LLM, \textsc{UniWeb}, is depicted in Figure~\ref{fig-model}. We first transform knowledge-intensive tasks into a unified text-to-text paradigm and consider the web as a general form of knowledge source. Based on the retrieved knowledge, we further design two training objectives to build  our model. In the next sections, we will describe our method in detail.

\subsection{Knowledge-Intensive Tasks Unification}
\label{sec-unification}

Previous retrieval-augmented approaches  usually adopt diverse architectures and different types of knowledge resources~\cite{kit_survey}. 
Instead, we aim to leverage the general knowledge source (\ie the web) to develop a unified framework that can fulfill  various (or most) knowledge-intensive tasks. Specifically, we unify 16 typical knowledge-intensive tasks across 7 task families, including fact checking, slot filling, dialogue, open-domain question answering, commonsense question answering, commonsense reasoning, and natural language inference. We convert these tasks as a general text-to-text transformation for training a unified LLM.

These tasks are mainly from the studies~\cite{kilt,sphere}, in which the original tasks of fact checking, slot filling, dialogue, and open-domain QA are  designed specifically based on the retrieved  knowledge from Wikipedia, while other tasks of commonsense QA, commonsense reasoning, and natural language inference focus on some more specific commonsense knowledge, going beyond Wikipedia. We consider these knowledge-intensive tasks as typical NLP tasks to show that the large-scale web can be specially useful for satisfying diverse information needs. More details about each task can be found in Appendix~\ref{app-dataset}.


\subsection{Web-based Knowledge Retrieval}
\label{sec-retrieval}

Unlike prior work that retrieves documents from offline corpora such as Wikipedia~\cite{realm,rag}, we propose to retrieve  \textit{comprehensive} and \emph{up-to-date} information from the online web through a general-purpose search engine. 
Although it is intuitive to extend the retrieval-augmented framework with the web as the knowledge resource, it is non-trivial to effectively leverage the knowledge found on the web. 
The documents on the web have inconsistent quality, and  contain noisy, biased, or even harmful contents~\cite{luccioni2021s}. Low-quality content may steer LLMs towards seemingly plausible but factually incorrect outputs~\cite{ji2022survey}.  
On the other hand, compared to a local neural retriever, black-box search engines can only be accessed through queries, which is less controllable and not easy to filter out noisy contents from the search results. 
In addition, LLMs do not always need external knowledge for task solving, especially for easy tasks. Therefore, we should request for more knowledge only when needed.


\subsubsection{LLM Knowledge Evaluation} 
\label{sec-self-evaluation}



To address the above challenges, it is essential to evaluate 
LLMs' own capabilities in a task and the necessity to refer to external knowledge. In our approach, we  consider a non-trivial question before retrieval: \textit{does a LLM need to retrieve knowledge for a specific task instance?}
For this purpose, we investigate whether or not LLMs can correctly answer questions without using external evidence. According to the recent study~\cite{kadavath2022language}, LLMs can self-evaluate the confidence level of their generation results (\eg \emph{True} or \emph{False}). 
Hence, we propose to utilize the self-evaluation mechanism 
to determine whether it is necessary to access additional web information. 

\paratitle{Self-Evaluation.} Specifically, we hypothesize that when a model ``knows'' the true output (\ie confident about its output) for a specific input,  sampling the outputs many times would result in an output distribution with small entropy. Following \citet{kadavath2022language}, we sample $n$ ($n$ = 200) different outputs for each input and estimate the entropy of the output distribution as follows:
\begin{eqnarray}\label{eq-entropy}
    H(\hat{\mathcal{Y}}|\mathcal{X}) &=& \mathbb{E}_{\hat{\mathcal{Y}}\sim\mathcal{G}}[-\log \text{Pr}(\hat{\mathcal{Y}}|\mathcal{X})] \\
    &=& \mathbb{E}_{\hat{\mathcal{Y}}\sim\mathcal{G}}\left[-\sum_{w_i \in \hat{\mathcal{Y}}} \log \text{Pr}(w_i|\mathcal{X},w_{<i})\right], \nonumber
\end{eqnarray}
where $\hat{\mathcal{Y}}=\langle w_1,...,w_i,...,w_m \rangle$ is the output text generated by the model $\mathcal{G}$. Then, we set an entropy threshold $\eta$. If $H(\hat{\mathcal{Y}}|\mathcal{X})$ is higher than $\eta$, it means that the model is unconfident about its outputs and needs supporting evidence from the web, otherwise, it does not. 
We will further demonstrate the predictive power of the entropy (Eq.~\eqref{eq-entropy}) in estimating the model confidence for knowledge retrieval. 


\subsubsection{Web Knowledge Retrieval} 

In \emph{active learning}~\cite{ren2021survey}, a prediction model can interactively query for labeling examples with low confidence levels. This learning method can not only reduce the cost of data labeling, but also remove those noisy and unhelpful data that models cannot benefit from. Inspired by this, we propose a \textit{search engine assisted learning} approach, in which LLMs choose those hard cases that they cannot solve  (assessed by self-evaluation) to query the off-the-shelf search engine for knowledge retrieval.
Different from active learning, our approach does not directly query for the final answer (largely reducing the labeling efforts), but instead the supporting evidence for solving the task. 
After retrieving knowledge from the web, it is critical to filter out noisy contents and select the most helpful and relevant knowledge that can enhance LLMs' confidence to generate correct outputs. Therefore, we elaborate a \emph{two-stage filter mechanism} to filter the retrieved knowledge.


\paratitle{Search Engine  Assisted Learning.} Specifically, for those hard examples, we take their input text $\mathcal{X}$ verbatim as a search query and issue a call to Google Search via API. For each query, we retrieve top-$K$ HTML pages and parse them to obtain clean texts, resulting in a set of passages $\mathcal{P}=\{p_1,...,p_K\}$. To filter out  noisy and irrelevant information, in the first stage, we chunk each passage into paragraphs, compute the cosine similarity between input and paragraph embeddings, and select the five most relevant paragraphs to form the final passage. In the second stage, we adopt the same method as self-evaluation (Eq.~\ref{eq-entropy}) to compute the model confidence given the input and each processed passage and select those passages  with high confidence as the final evidence.



\subsection{Knowledge-based Model Pre-training}
\label{sec-pretrain}

In most previous work, the retrieval model is either pretrained using self-supervised objective such as MLM~\cite{realm,borgeaud2022improving} or trained for specific tasks~\cite{rag}. In this work, we focus on explicitly training web-augmented LLMs in a supervised and massively multi-task fashion~\cite{ext5} using the mixture of knowledge-intensive tasks (Section~\ref{sec-unification}). Besides, to integrate the retrieved knowledge into LLMs, we design a continual knowledge learning task based on the retrieved passages. 


\paratitle{Knowledge-Intensive Learning.} This 
pretraining objective uses the retrieved knowledge and labeled data from the unified knowledge-intensive tasks. Formally, given an input text $\mathcal{X}$ and retrieved passages $\mathcal{P}$, this objective is to minimize the negative log-likelihood loss over the output text $\mathcal{Y}$:
\begin{equation}\label{eq-kil}
    \mathcal{L}_{KIL} = - \sum_{i=1}^{m} \log \text{Pr}(w_i|w_{<i},\mathcal{X},\mathcal{P}),
\end{equation}
where $w_i$ denotes the $i$-th token of the output text $\mathcal{Y}$. We concatenate the input text $\mathcal{X}$ and retrieved passages $\mathcal{P}$ using the manually-written task-specific prompts (shown in Appendix~\ref{app-dataset}). 
Pretrained on the unified knowledge-based text-to-text format, our model can be easily applied to diverse knowledge-intensive tasks. It has been reported that ensembling many tasks, distributions and domains during pretraining can improve LLMs' generalization to new tasks~\cite{ext5}.

\paratitle{Continual Knowledge Learning.} Due to the limited pretraining on single static corpus, the knowledge encoded in LLMs has a discrepancy with the retrieved knowledge from the web. Thus, to reduce the discrepancy and integrate the newly retrieved knowledge into LLMs, we design a self-supervised pretraining task, \ie continual knowledge learning. For most knowledge-intensive tasks such as slot filling and fact verification, named entities are of special importance. 
Thus, 
this pretraining task aims to predict the salient masked spans (\ie named entities) in retrieved passages.
Firstly, we use a BERT-based~\cite{bert} tagger trained on CoNLL-2003 data~\cite{sang2003introduction} to identify name entities and then mask entities such as ``United States''. Then, our model will be trained to predict these masked spans by minimizing the masked span prediction loss:
\begin{equation}\label{eq-ckl}
    \mathcal{L}_{CKL} = - \sum_{k=1}^{K} \sum_{j=1}^{m} \log \text{Pr}(s_j|\tilde{p}_k),
\end{equation}
where $s_j$ is the $j$-th masked span for the passage $p_k$, and $\tilde{p}_k$ denotes the unmasked tokens in $p_k$.

\section{Experiments}

In this section, we detail the experimental setup and then highlight the main observations of our results.

\subsection{Experimental Setup}
\paratitle{Knowledge Source.} 
In large-scale pre-training, we leverage an open massive web corpus CCNet~\cite{ccnet} to provide documents with diverse topics, approximating the realistic web. Following \citet{sphere}, we select the CCNet snapshot corresponding to the August 2019 Common Crawl snapshot which covers a wide range of 134M web documents and finally yields 906M passages of 100 tokens. CCNet processes Common Crawl through deduplication, language identification and quality filtering based on perplexity calculated by a language model. In downstream fine-tuning, we test with the off-the-shelf search engine, \ie Google Search, to retrieve documents from the real-time web. Specifically, we utilize the input text verbatim as query and request a call to Google Search via API\footnote{https://developers.google.com/custom-search}. Besides, for the Wikipedia-based baselines, we use the 2019/08/01 Wikipedia snapshot from the KILT benchmark~\cite{kilt}, consisting of 5.9M documents split into 22.2M passages of 100 tokens. This data snapshot is temporally the closest to the CCNet corpus for fair comparison.


\begin{table*}[t]
	\small
	\centering
	\begin{tabular}{r c c c c c c c c}
		\toprule
		\multicolumn{1}{c}{\multirow{2.5}{*}{\textbf{Models}}} & Fact Checking & \multicolumn{2}{c}{Slot Filling} & Dialogue & \multicolumn{4}{c}{Open-domain QA} \\ 
		\cmidrule(r){2-2}\cmidrule(r){3-4}\cmidrule(r){5-5}\cmidrule(r){6-9}
		& \textbf{FEVER}  & \textbf{T-REx} & \textbf{zsRE} & \textbf{WoW} & \textbf{NQ} & \textbf{HotpotQA} & \textbf{TriviaQA} & \textbf{ELI5} \\ 
            \midrule[0.5pt]
            \multicolumn{9}{c}{\textit{w/o Retrieval}} \\
            \specialrule{0em}{1pt}{1pt}
            \textbf{BART}$_\textsc{Large}$ & 78.93 & 45.06 & \;\,9.14 & 12.86 &  21.75 & 15.37 & 32.39 & \uline{20.55} \\
            \textbf{T5}$_\textsc{Large}$ & 80.31 & 50.63 & 10.34 & 12.67 & 28.50 & 18.98 & 35.90 & \textbf{20.60} \\
            \midrule[0.5pt]
            \multicolumn{9}{c}{\textit{w/ Wikipedia}} \\
            \specialrule{0em}{1pt}{1pt}
            \textbf{REALM} &  76.22 & 53.35 & 39.38 & - & 40.40 & 22.23 & 65.44 & 10.23 \\
            \textbf{RAG} &  86.31 & 59.20 & 44.74 & 13.11 & 44.39 & 26.97 & 71.27 & 14.05 \\
            \textbf{BART+DPR} & 86.74 &  59.16 & 30.43 &  15.19 & 41.27 & 25.18 & 58.55 & 17.41 \\
            \textbf{BART+DPR}$_\textsc{Multi}$ & 86.32 & 78.50 & 57.95 & 15.33 & 39.75 & 31.77 & 59.60 & 17.07 \\
            \textbf{FID+DPR}$_\textsc{Multi}$ & 88.99 &  \uline{82.19} & \uline{71.53} & 15.66 &  \uline{49.86} & \uline{36.90} & 71.04 & 16.45 \\
            \multicolumn{9}{c}{\textit{w/ CCNet}} \\
            \specialrule{0em}{1pt}{1pt}
            \textbf{FID+DPR}$_\textsc{Multi}$ & 85.74 & 52.06 & 28.47 & 15.22 & 45.15 & 27.29 & 67.49 & 16.14 \\
            \textbf{FID+DPR}$_\textsc{CCNet}$ & 87.43 & 57.02 & 36.55 & 15.29 & 48.61 & 31.64 & 73.06 & 15.76 \\
            \textbf{FID+BM25} & \uline{89.12} & 62.12 & 43.92 & \uline{17.28} & 46.05 & 34.10 & \textbf{78.21} & 15.59 \\
            \midrule[0.5pt]
            \multicolumn{9}{c}{\textit{w/ Web}} \\
            \textbf{UniWeb} & \textbf{91.69} & \textbf{83.58} & \textbf{72.42} & \textbf{20.87} & \textbf{54.37} & \textbf{40.73} & \uline{77.01} & 18.34 \\
            \bottomrule
	\end{tabular}
	\caption{Evaluation results on the test set for fact checking, slot filling, dialogue, and open-domain QA. We report \textit{Accuracy} for FEVER, T-REx, and zsRE; \textit{EM} for NQ, HotpotQA, and TriviaQA; \textit{ROUGE-L} for ELI5 and \textit{F1-score} for WoW. These results come from no-retrieval models (top section), Wikipedia/CCNet-based models (middle section), and Web-based models (bottom section). \textbf{Bold} and \underline{underline} denote the best and second best methods. }
        \vspace{-0.1cm}
	\label{tab:main-first}
\end{table*}

\paratitle{Pre-training Tasks.} As described in Section~\ref{sec-unification}, we unify 16 knowledge-intensive tasks across seven task families during pre-training:

\begin{itemize}[leftmargin=*,itemsep=-1.5pt,topsep=4pt]
    \item \textbf{Fact Checking}: FEVER~\cite{fever}.
    \item \textbf{Slot Filling}: T-REx~\cite{trex} and zero-shot RE~\cite{zsre}.
    \item \textbf{Dialogue}: Wizard-of-Wikipedia~\cite{wow}.
    \item \textbf{Open-domain QA}: TriviaQA~\cite{triviaqa}, Natural Questions~\cite{nq}, HotpotQA~\cite{hotpotqa}, and ELI5~\cite{eli5}.
    \item \textbf{Commonsense QA}: CommonsenseQA~\cite{csqa}, SocialIQa~\cite{socialiqa}, CosmosQA~\cite{cosmosqa}, and PIQA~\cite{piqa}.
    \item \textbf{Commonsense Reasoning}: NumerSense~\cite{numersense} and WinoGrande~\cite{winogrande}.
    \item \textbf{Natural Language Inference}: $\alpha$NLI~\cite{anli} and HellaSwag~\cite{hellaswag}.
\end{itemize}

We convert these tasks into a unified text-to-text format. We take the input text as query to retrieve top $10$ passages from CCNet. After pre-processing, we mix the training set of these datasets to pre-train our model. 
We present the statistics of datasets and pre-processing details in Appendix~\ref{app-dataset}.


\paratitle{Baselines}. We compare \textbf{UniWeb} to a wide range of models as follows:

\begin{itemize}[leftmargin=*,itemsep=-1.5pt,topsep=4pt]
    \item \textbf{BART}~\cite{bart} and \textbf{T5}~\cite{t5}. These are two representative text-to-text LLMs for solving knowledge-intensive tasks. We adopt the large version for a fair comparison.
    \item \textbf{REALM}~\cite{realm} and \textbf{RAG}~\cite{rag}. They are two well-known retrieval-augmented LLMs combining with a nonparametric memory of Wikipedia via a neural retriever.
    \item \textbf{Fusion-in-Decoder (FID)}~\cite{fid}. It is based on T5 where the encoder encodes the input text with each passage and the decoder combines the encoded representations.
    \item \citet{maillard2021multi} and \citet{sphere} equip BART and FID with retrieval models, \ie \textbf{BM25}~\cite{bm25}, \textbf{DPR}~\cite{dpr}, \textbf{DPR$_\textsc{Multi}$} trained in a multi-task fashion, and \textbf{DPR$_\textsc{CCNet}$} trained on CCNet.
\end{itemize}

Note that these models are trained on individual tasks and datasets, while our model is pre-trained in a multi-task manner. We use BM25 to retrieve passages from CCNet during pretraining.
The BM25 and DPR indices are collected from the previous word~\cite{sphere}. Since it lacks the retrieval supervision to train DPR for those tasks in Table~\ref{tab:main-second}, we only report the BM25 results. The implementation details are shown in Appendix~\ref{app-configuration}.

\begin{table*}[t]
\renewcommand\arraystretch{1.1}
\setlength\tabcolsep{5.3pt}
	\small
	\centering
	\begin{tabular}{r c c c c c c c c}
		\toprule
		\multicolumn{1}{c}{\multirow{2.5}{*}{\textbf{Models}}} & \multicolumn{4}{c}{Commonsense QA} & \multicolumn{2}{c}{Commonsense Reasoning} & \multicolumn{2}{c}{NLI} \\ 
            \cmidrule(r){2-5}\cmidrule(r){6-7}\cmidrule(r){8-9}
		& \textbf{CSQA}  & \textbf{SocialIQA} & \textbf{CosmosQA} & \textbf{PIQA} & \textbf{NumerSense} & \textbf{WinoGrande} & \textbf{HellaSwag} & \textbf{$\alpha$NLI} \\ 
            \midrule[0.5pt]
            \multicolumn{9}{c}{\textit{w/o Retrieval}} \\
            \specialrule{0em}{1pt}{1pt}
            \textbf{BART}$_\textsc{Large}$ & 62.50 & 74.00 & 75.11 & 77.40 & 55.30 & 62.40 & 76.60 & 75.12 \\
            \textbf{T5}$_\textsc{Large}$ & 72.56 & \uline{74.16} & 79.23 & 78.67 & 59.71 & 76.48 & 79.84 & 77.48 \\
            \midrule[0.5pt]
            \multicolumn{9}{c}{\textit{w/ Wikipedia}} \\
            \specialrule{0em}{1pt}{1pt}
            \textbf{REALM} & 63.11 & 62.52 & 71.33 & 70.65 & 57.34 & 62.12 & 73.21 & 71.40 \\
            \textbf{RAG} & 69.51 & 68.32 & 76.55 & 75.23 & 59.22 & 63.35 & 75.01 & 74.45 \\
            \textbf{BART+BM25} & 70.16 & 70.83 & 76.14 & 77.04 & 57.50 & 65.09 & 76.34 & 74.66 \\
            \textbf{FID+BM25} & \uline{73.63} & \textbf{74.36} & 78.83 & 79.65 & 62.30 & 76.72 & 79.96 & \textbf{77.94} \\
            \multicolumn{9}{c}{\textit{w/ CCNet}} \\
            \specialrule{0em}{1pt}{1pt}
            \textbf{FID+BM25} & \uline{73.63} & 73.64 & \uline{79.63} & \textbf{81.66} & \uline{66.70} & \uline{76.80} &  \uline{81.96} & \uline{77.74} \\
            \midrule[0.5pt]
            \multicolumn{9}{c}{\textit{w/ Web}} \\
            \textbf{UniWeb} & \textbf{75.34} & 73.17 & \textbf{80.96} & \uline{79.77} & \textbf{69.23} & \textbf{78.74} & \textbf{82.12} & 77.23 \\
            \bottomrule
	\end{tabular}
	\caption{Evaluation results at \textit{Accuracy} on the dev set for commonsense QA, commonsense reasoning, and natural language inference (NLI). \textbf{Bold} and \underline{underline} numbers denote the best and second best performance. Following \citet{sphere}, since it lacks the retrieval supervision to train DPR, we only report the BM25 results.}
	\label{tab:main-second}
\end{table*}

\paratitle{Evaluation Metrics}. We adopt various tasks and datasets in our experiments, which need to be evaluated differently. Following~\citet{kilt}, we use \textit{Exact Match} (EM) for datasets with extractive (\ie Natural Questions, TriviaQA) or short abstractive output text (\ie HotpotQA); for datasets with long abstractive output text, we use \textit{ROUGE-L} \cite{lin2004rouge} for ELI5 and \textit{F1-score} for Wizard of Wikipedia; we use \textit{Accuracy} for the remaining tasks. To compute EM and F1-score, we conduct post-processing on the gold and predicted output texts such as lowercasing, stripping, punctuation, and duplicate whitespace~\cite{squad}. 

\subsection{Main Results}
\label{sec-full}

Table~\ref{tab:main-first} and Table~\ref{tab:main-second} show the results of \textsc{UniWeb} and baselines on 16 knowledge-intensive tasks.

First, on almost all knowledge-intensive tasks, combining LLMs with explicit retrieved knowledge can achieve higher performance. From Wikipedia and CCNet to the web, we can observe that a broader coverage of knowledge will lead to better results. Compared to BART and T5, retrieval-based models benefit from the retrieved knowledge.

Second, the tasks in Table~\ref{tab:main-first} are specially designed based on the knowledge from Wikipedia. Thus, there is a strong bias towards Wikipedia as the knowledge resource. 
We can observe that CCNet only achieves comparable results or even suffers from a large performance drop. However, for the tasks in Table~\ref{tab:main-second} requiring knowledge beyond Wikipedia, CCNet is more competitive.

Finally, our \textsc{UniWeb} model achieves the best results on most knowledge-intensive tasks. On one hand, our model is trained in a multi-task manner, which can benefit from knowledge sharing across tasks. On the other hand, our model can access broad and up-to-date knowledge from the web via the fine-tuned search engine. The web knowledge can fulfill more diverse information needs. Moreover, the search engine works much better than traditional sub-optimal retrieval methods that rely on end-to-end training or word matching.

\begin{table}[t]
	\small
	\centering
	\setlength\tabcolsep{3.5pt}
	\begin{tabular}{lrrrrr}
		\toprule
		 \textbf{Models} & \textbf{zsRE} & \textbf{WoW} & \textbf{CSQA} & \textbf{PIQA} & \textbf{$\alpha$NLI} \\
		\midrule
		UniWeb    & 72.42 &  20.87  &  75.34 & 79.77 & 77.23 \\
		\midrule
            w/ Wikipedia & 70.23  & 16.34 & 62.77 & 77.45 & 74.46 \\
            w/ CCNet & 43.25 & 17.23 & 70.89 & 79.45 & 76.01 \\
		w/o SE  & 68.34     & 19.17     & 67.44 & 76.80 & 73.90  \\
            w/o CKL & 69.70 & 19.09 & 66.70 & 76.57 & 75.01 \\
		\bottomrule
	\end{tabular}
	\caption{Ablation study on five tasks.}
        \vspace{-0.3cm}
	\label{tab:ablation}
\end{table}

\subsection{Detailed Analysis} 
\label{sec-detailed-exp}

We report detailed analysis of UniWeb in several datasets -- we have similar finding in other datasets. 


\paratitle{Ablation Study.} Our \textsc{UniWeb} model is the first unified LLM using the web as knowledge source for knowledge-intensive tasks. To examine the importance of the web, we design two counterparts: (1) \textit{w/ Wikipedia} or (2) \textit{w/ CCNet} replaces the web with Wikipedia or CCNet and adopts BM25 to retrieve documents. Besides, to avoid the negative impact of noisy and biased information, we adopt the self-evaluation method to adaptively access knowledge from the web. Thus, we remove this method to test its effect (\textit{w/o SE}). Finally, we remove the pretraining task, \ie continuous knowledge learning, to test its importance (\textit{w/o CKL}). 
The results are shown in Table~\ref{tab:ablation}. 
We can see that replacing the web with Wikipedia or CCNet suffers from a large performance drop.
Besides, the self-evaluation method benefits our model a lot in terms of knowledge filtering. The pretraining task also improves the knowledge capacity of our model.

\begin{table*}[t]
	\small
	\centering
	\begin{tabular}{p{0.31\textwidth} p{0.31\textwidth} p{0.31\textwidth}}
		\toprule
		  \multicolumn{3}{l}{\tabincell{l}{\textbf{Question:} With France and Argentina set to battle it out on Sunday in the \textbf{World Cup final 2022}, which teams will go head \\to head for \textbf{the third place}?}} \\
            \multicolumn{3}{l}{\textbf{Gold Answer:} Croatia and Morocco} \\
		\midrule
            Top-1 Wikipedia Passage & Top-1 CCNet Passage & Top-1 Web Passage \\
            \midrule
            ...~\textcolor{red}{Third place} play-off \textbf{The Netherlands} \textcolor{red}{defeated} \textbf{Brazil} 3–0 to \textcolor{red}{secure third place}, the first for the Dutch team in their history. Overall, Brazil conceded 14 goals in the tournament; this was the most by a team at any single \textcolor{red}{World Cup} since 1986, and the most by a host nation in history... \makecell[l]{ \url{https://en.wikipedia.org}\\\url{/wiki/2014_FIFA_World_Cup}}
            &
            ...~\textbf{France} and \textbf{Belgium} \textcolor{red}{go head-to-head} in the first semi-finals of \textcolor{red}{World Cup 2018}. Both teams have impressed in Russia so far, but only one can make it through to Sunday's final. However, Les Bleus have won four of their five matches at  \textcolor{red}{World Cup 2018} and shown flashes of quality in the process... \makecell[l]{ \url{https://myarsenalblog.com}\\\url{/category/uncategorized}}
            &
            ...~\textcolor{red}{Third place} for Croatia Zlatko Dalic's \textbf{Croatia} followed up their runners-up effort at the Russia 2018 World Cup with \textcolor{red}{third place in Qatar} as Mislav Orsic’s fine effort \textcolor{red}{secured victory over} the tournament's surprise package \textbf{Morocco} at Khalifa International Stadium... \makecell[l]{ \url{https://ca.sports.yahoo.com}\\\url{/news/today-world-cup-argen}\\\url{tina-head-085045315.html}} \\
            \midrule 
            \textbf{Prediction:} The Netherlands and Brazil & \textbf{Prediction:} France and Belgium & 
            \textbf{Prediction:} Croatia and Morocco \textcolor{red}{\CheckmarkBold} \\
            \bottomrule
	\end{tabular}
	\caption{A qualitative example showing the top-1 retrieved passages from Wikipedia, CCNet, and web, and their corresponding model prediction. The words in red denote the keywords related to the question.}
	\label{tab:case-study}
\end{table*}

\begin{figure}[t]
	\centering
	\subfigure[]{
		\centering
		\includegraphics[width=0.235\textwidth]{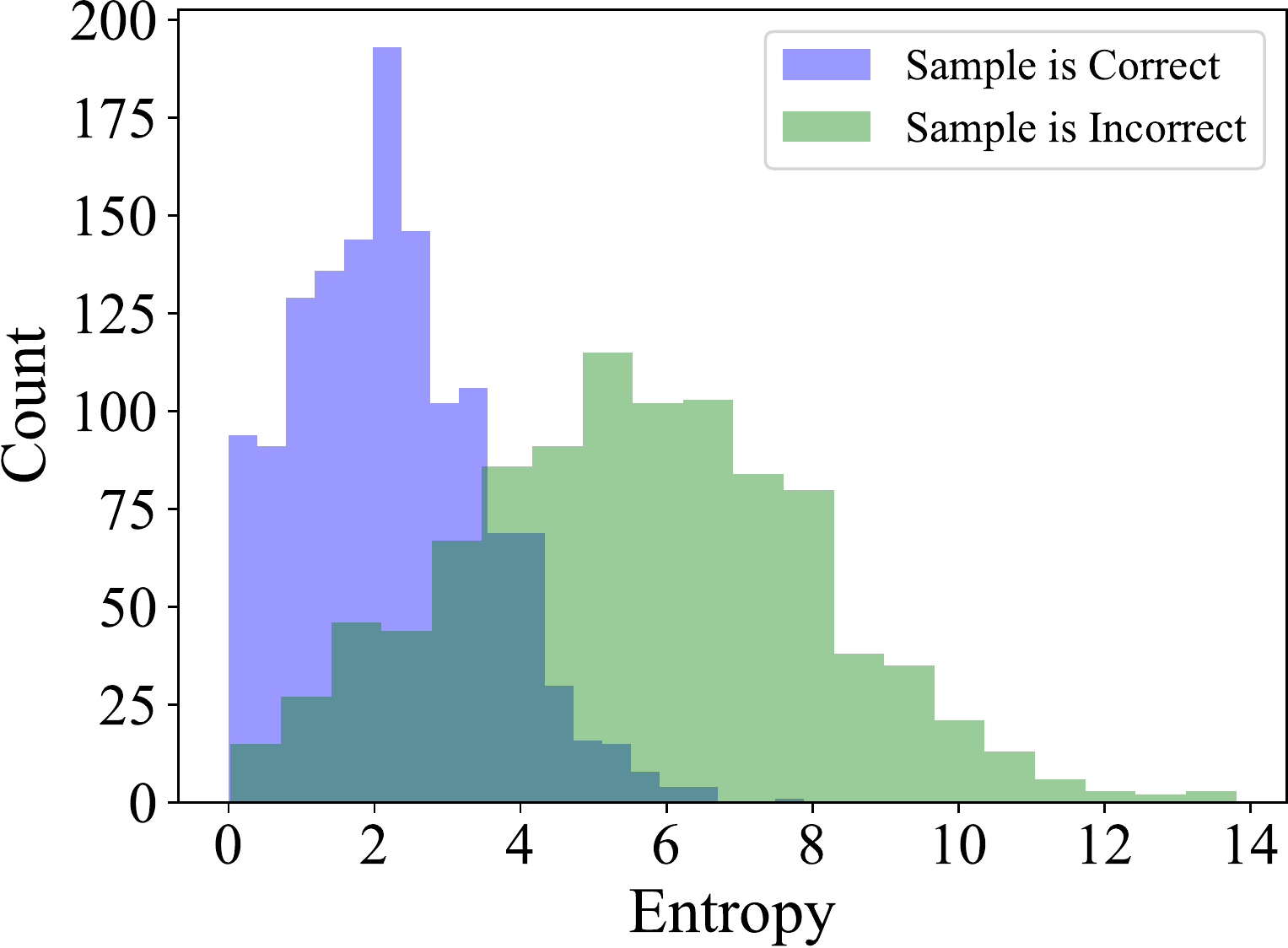}
	}
	\subfigure[]{
		\centering
		\includegraphics[width=0.21\textwidth]{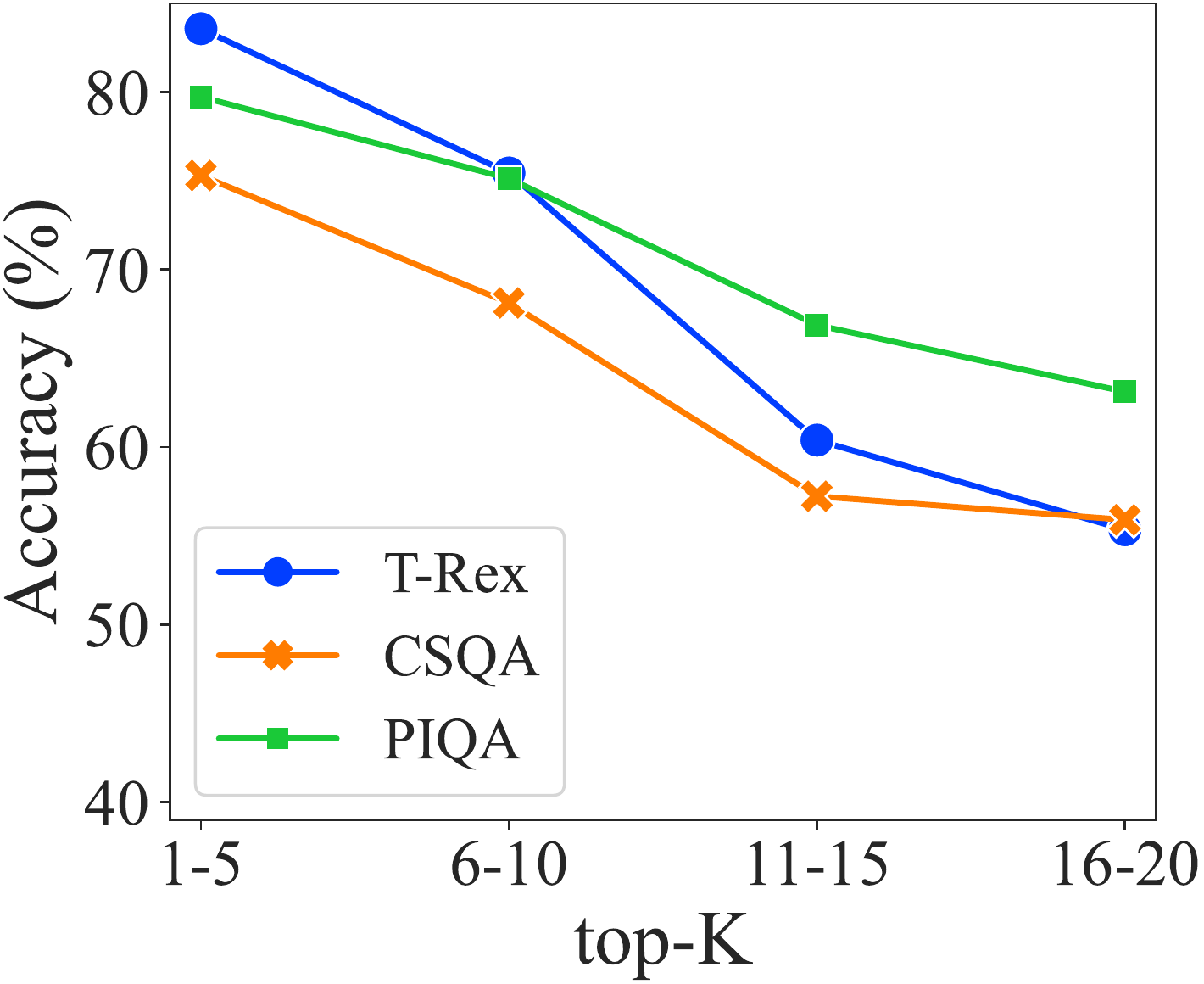}
	}
	\centering
	\caption{(a) Entropy of samples in HotpotQA; (b) Accuracy \textit{w.r.t} different top-$K$ documents.}
	\label{fig:sensitivity}
	\vspace{-0.2cm}
\end{figure}

\paratitle{Sensitivity Analysis.} In the self-evaluation mechanism, we use entropy to evaluate the model confidence. 
To verify its effectiveness, we present the distribution of $H(\mathcal{\hat{Y}}|\mathcal{X})$ depending on whether or not the model gets the question correct. As shown in Figure~\ref{fig:sensitivity}(a), the average entropy of the questions for which our model gets correct is lower than that of questions for which our model gets incorrect. This indicates that the entropy has some predictive power of model confidence. Besides, the quality of retrieved documents will largely affect the prediction of our model. Thus, in Figure~\ref{fig:sensitivity}(b), we test the model accuracy by varying the top-$K$ search results in the set of \{1-5, 6-10, 11-15, 16-20\}. We can see that LLM performance drops with the increase of rank of documents, thus the decrease of document quality. However, the retrieved top 6-10 passages also achieve comparable results to the top 1-5 ones. This is the motivation of our setting $K=10$.





\subsection{Case Study} 

In this section, we perform the qualitative analysis on \textsc{RealTime QA}~\cite{realtimeQA}, a benchmark requiring real-time, up-to-date, and comprehensive knowledge with a broad range of topics (such as politics, business, sports, and entertainment) to solve questions. The evaluation results are shown in Appendix~\ref{app-experiment}. Our UniWeb model with Google Search performs the best. We present an example in Table~\ref{tab:case-study} about ``World Cup final 2022'' in the sports topic. By using the question text as query, we can retrieve top-1 passages from Wikipedia, CCNet, and web. Since Wikipedia and CCNet are both static and limited knowledge resources, the retrieved passages are not fresh 
in time (``2014'' and ``2018'') even though they are on the same topic ``World Cup''. The typical retrieval methods (BM25 or DPR) are largely reliant on fuzzy semantic matching, also leading to incorrect retrieval. While, retrieving from the web using search engine can ensure our model to obtain the most up-to-date and relevant information, based on which it can generate the correct answer ``Croatia and Morocco''. We present more examples in Appendix~\ref{app-examples}.

\section{Conclusion}

This paper presented a unified web-augmented framework for a wide range of knowledge-intensive tasks, called \textsc{UniWeb}. We convert 16 tasks into a text-to-text generation task for training. We propose a search engine assisted learning method to selectively retrieve documents from the web through Google Search. Furthermore, to reduce the discrepancy between the encoded and retrieved knowledge, we design a pre-training task, \ie continual knowledge learning, to integrate the retrieved knowledge into LLMs. Experiments on 16 tasks show the effectiveness of our web-augmented model compared to previous retrieval-augmented models. In future work, we will investigate the effect of web content in detail and consider applying our model to more types of downstream tasks.

\section{Limitations}

For web-augmented models including our work, the deterioration of search results from search engine highlights the importance of deriving an effective method to interact with the huge web. 
Search engines are often perceived as black-box and non-transparent for end users. Therefore, many works proposed ``leaning to search'' to decompose complex questions into simpler queries, which may improve the performance of web-based models~\cite{webgpt,komeili2021internet}.

In our model, we used a commercial search engine as the retriever to work with the whole web as a knowledge source. Since the web is not  curated and well-structured like Wikipedia, we may encounter unexpected safety issues, including misinformation and harmful contents. While we have relied on the security control of the search engine, 
more attention should be paid to better understand the risks and provide effective ways to mitigate them. We hope our simple approach and strong results could encourage more future work by the community to tackle these questions. To encourage the community to investigate the question and ensure reproducibility, after the reviewing process, we will release the search URLs used in our experiments.

As for the potential concern, since we use the search engine to access real-time information, we do not have a tight control over retrieved results as traditional end-to-end retrieval~\cite{realm,rag}. Not only the changes of search engine logic, but also the newly published information, might create  discrepancies over the course of time. This is also an issue we have to tackle to build a stable web-based solution for LLMs.

\section*{Acknowledgments}

This work was partially supported by National Natural Science Foundation of China under Grant No. 62222215, Beijing Natural Science Foundation under Grant No. 4222027, Beijing Outstanding Young Scientist Program under Grant No. BJJWZYJH012019100020098, and the Outstanding Innovative Talents Cultivation Funded Programs 2021 of Renmin University of China.  Xin Zhao is the corresponding author.

\bibliography{anthology}

\begin{thebibliography}{67}
\expandafter\ifx\csname natexlab\endcsname\relax\def\natexlab#1{#1}\fi

\bibitem[{Aribandi et~al.(2022)Aribandi, Tay, Schuster, Rao, Zheng, Mehta,
  Zhuang, Tran, Bahri, Ni, Gupta, Hui, Ruder, and Metzler}]{ext5}
Vamsi Aribandi, Yi~Tay, Tal Schuster, Jinfeng Rao, Huaixiu~Steven Zheng,
  Sanket~Vaibhav Mehta, Honglei Zhuang, Vinh~Q. Tran, Dara Bahri, Jianmo Ni,
  Jai Gupta, Kai Hui, Sebastian Ruder, and Donald Metzler. 2022.
\newblock \href {https://openreview.net/forum?id=Vzh1BFUCiIX} {Ext5: Towards
  extreme multi-task scaling for transfer learning}.
\newblock In \emph{International Conference on Learning Representations}.

\bibitem[{Bhagavatula et~al.(2020)Bhagavatula, Bras, Malaviya, Sakaguchi,
  Holtzman, Rashkin, Downey, Yih, and Choi}]{anli}
Chandra Bhagavatula, Ronan~Le Bras, Chaitanya Malaviya, Keisuke Sakaguchi, Ari
  Holtzman, Hannah Rashkin, Doug Downey, Wen{-}tau Yih, and Yejin Choi. 2020.
\newblock Abductive commonsense reasoning.
\newblock In \emph{8th International Conference on Learning Representations,
  {ICLR} 2020, Addis Ababa, Ethiopia, April 26-30, 2020}. OpenReview.net.

\bibitem[{Bisk et~al.(2020)Bisk, Zellers, LeBras, Gao, and Choi}]{piqa}
Yonatan Bisk, Rowan Zellers, Ronan LeBras, Jianfeng Gao, and Yejin Choi. 2020.
\newblock {PIQA:} reasoning about physical commonsense in natural language.
\newblock In \emph{The Thirty-Fourth {AAAI} Conference on Artificial
  Intelligence, {AAAI} 2020, The Thirty-Second Innovative Applications of
  Artificial Intelligence Conference, {IAAI} 2020, The Tenth {AAAI} Symposium
  on Educational Advances in Artificial Intelligence, {EAAI} 2020, New York,
  NY, USA, February 7-12, 2020}, pages 7432--7439. {AAAI} Press.

\bibitem[{Borgeaud et~al.(2022)Borgeaud, Mensch, Hoffmann, Cai, Rutherford,
  Millican, Van Den~Driessche, Lespiau, Damoc, Clark
  et~al.}]{borgeaud2022improving}
Sebastian Borgeaud, Arthur Mensch, Jordan Hoffmann, Trevor Cai, Eliza
  Rutherford, Katie Millican, George~Bm Van Den~Driessche, Jean-Baptiste
  Lespiau, Bogdan Damoc, Aidan Clark, et~al. 2022.
\newblock Improving language models by retrieving from trillions of tokens.
\newblock In \emph{International conference on machine learning}, pages
  2206--2240. PMLR.

\bibitem[{Brown et~al.(2020)Brown, Mann, Ryder, Subbiah, Kaplan, Dhariwal,
  Neelakantan, Shyam, Sastry, Askell et~al.}]{gpt3}
Tom~B Brown, Benjamin Mann, Nick Ryder, Melanie Subbiah, Jared Kaplan, Prafulla
  Dhariwal, Arvind Neelakantan, Pranav Shyam, Girish Sastry, Amanda Askell,
  et~al. 2020.
\newblock Language models are few-shot learners.
\newblock \emph{arXiv preprint arXiv:2005.14165}.

\bibitem[{Chen and Yih(2020)}]{ChenY20}
Danqi Chen and Wen{-}tau Yih. 2020.
\newblock Open-domain question answering.
\newblock In \emph{Proceedings of the 58th Annual Meeting of the Association
  for Computational Linguistics: Tutorial Abstracts, {ACL} 2020, Online, July
  5, 2020}, pages 34--37. Association for Computational Linguistics.

\bibitem[{Chowdhery et~al.(2022)Chowdhery, Narang, Devlin, Bosma, Mishra,
  Roberts, Barham, Chung, Sutton, Gehrmann et~al.}]{chowdhery2022palm}
Aakanksha Chowdhery, Sharan Narang, Jacob Devlin, Maarten Bosma, Gaurav Mishra,
  Adam Roberts, Paul Barham, Hyung~Won Chung, Charles Sutton, Sebastian
  Gehrmann, et~al. 2022.
\newblock Palm: Scaling language modeling with pathways.
\newblock \emph{arXiv preprint arXiv:2204.02311}.

\bibitem[{Devlin et~al.(2019)Devlin, Chang, Lee, and Toutanova}]{bert}
Jacob Devlin, Ming{-}Wei Chang, Kenton Lee, and Kristina Toutanova. 2019.
\newblock \href {https://doi.org/10.18653/v1/n19-1423} {{BERT:} pre-training of
  deep bidirectional transformers for language understanding}.
\newblock In \emph{Proceedings of the 2019 Conference of the North American
  Chapter of the Association for Computational Linguistics: Human Language
  Technologies, {NAACL-HLT} 2019, Minneapolis, MN, USA, June 2-7, 2019, Volume
  1 (Long and Short Papers)}, pages 4171--4186. Association for Computational
  Linguistics.

\bibitem[{Dinan et~al.(2019)Dinan, Roller, Shuster, Fan, Auli, and
  Weston}]{wow}
Emily Dinan, Stephen Roller, Kurt Shuster, Angela Fan, Michael Auli, and Jason
  Weston. 2019.
\newblock {W}izard of {W}ikipedia: Knowledge-powered conversational agents.
\newblock In \emph{Proceedings of the International Conference on Learning
  Representations (ICLR)}.

\bibitem[{Dong et~al.(2019)Dong, Yang, Wang, Wei, Liu, Wang, Gao, Zhou, and
  Hon}]{unilm}
Li~Dong, Nan Yang, Wenhui Wang, Furu Wei, Xiaodong Liu, Yu~Wang, Jianfeng Gao,
  Ming Zhou, and Hsiao-Wuen Hon. 2019.
\newblock \href
  {https://proceedings.neurips.cc/paper/2019/file/c20bb2d9a50d5ac1f713f8b34d9aac5a-Paper.pdf}
  {Unified language model pre-training for natural language understanding and
  generation}.
\newblock In \emph{Advances in Neural Information Processing Systems},
  volume~32. Curran Associates, Inc.

\bibitem[{Dou and Peng(2022)}]{dou2022zero}
Zi-Yi Dou and Nanyun Peng. 2022.
\newblock Zero-shot commonsense question answering with cloze translation and
  consistency optimization.
\newblock \emph{arXiv preprint arXiv:2201.00136}.

\bibitem[{ElSahar et~al.(2018)ElSahar, Vougiouklis, Remaci, Gravier, Hare,
  Laforest, and Simperl}]{trex}
Hady ElSahar, Pavlos Vougiouklis, Arslen Remaci, Christophe Gravier,
  Jonathon~S. Hare, Fr{\'{e}}d{\'{e}}rique Laforest, and Elena Simperl. 2018.
\newblock T-rex: {A} large scale alignment of natural language with knowledge
  base triples.
\newblock In \emph{Proceedings of the Eleventh International Conference on
  Language Resources and Evaluation, {LREC} 2018, Miyazaki, Japan, May 7-12,
  2018}. European Language Resources Association {(ELRA)}.

\bibitem[{Guo et~al.(2022)Guo, Schlichtkrull, and Vlachos}]{guo2022survey}
Zhijiang Guo, Michael Schlichtkrull, and Andreas Vlachos. 2022.
\newblock A survey on automated fact-checking.
\newblock \emph{Transactions of the Association for Computational Linguistics},
  10:178--206.

\bibitem[{Guu et~al.(2020)Guu, Lee, Tung, Pasupat, and Chang}]{realm}
Kelvin Guu, Kenton Lee, Zora Tung, Panupong Pasupat, and Mingwei Chang. 2020.
\newblock Retrieval augmented language model pre-training.
\newblock In \emph{International Conference on Machine Learning}, pages
  3929--3938. PMLR.

\bibitem[{Huang et~al.(2019)Huang, Bras, Bhagavatula, and Choi}]{cosmosqa}
Lifu Huang, Ronan~Le Bras, Chandra Bhagavatula, and Yejin Choi. 2019.
\newblock Cosmos {QA:} machine reading comprehension with contextual
  commonsense reasoning.
\newblock In \emph{Proceedings of the 2019 Conference on Empirical Methods in
  Natural Language Processing and the 9th International Joint Conference on
  Natural Language Processing, {EMNLP-IJCNLP} 2019, Hong Kong, China, November
  3-7, 2019}, pages 2391--2401. Association for Computational Linguistics.

\bibitem[{Huang et~al.(2020)Huang, Zhu, and Gao}]{huang2020challenges}
Minlie Huang, Xiaoyan Zhu, and Jianfeng Gao. 2020.
\newblock Challenges in building intelligent open-domain dialog systems.
\newblock \emph{ACM Transactions on Information Systems (TOIS)}, 38(3):1--32.

\bibitem[{Izacard and Grave(2020)}]{fid}
Gautier Izacard and Edouard Grave. 2020.
\newblock Leveraging passage retrieval with generative models for open domain
  question answering.
\newblock \emph{arXiv preprint arXiv:2007.01282}.

\bibitem[{Izacard et~al.(2022)Izacard, Lewis, Lomeli, Hosseini, Petroni,
  Schick, Dwivedi-Yu, Joulin, Riedel, and Grave}]{izacard2022few}
Gautier Izacard, Patrick Lewis, Maria Lomeli, Lucas Hosseini, Fabio Petroni,
  Timo Schick, Jane Dwivedi-Yu, Armand Joulin, Sebastian Riedel, and Edouard
  Grave. 2022.
\newblock Few-shot learning with retrieval augmented language models.
\newblock \emph{arXiv preprint arXiv:2208.03299}.

\bibitem[{Ji et~al.(2022)Ji, Lee, Frieske, Yu, Su, Xu, Ishii, Bang, Madotto,
  and Fung}]{ji2022survey}
Ziwei Ji, Nayeon Lee, Rita Frieske, Tiezheng Yu, Dan Su, Yan Xu, Etsuko Ishii,
  Yejin Bang, Andrea Madotto, and Pascale Fung. 2022.
\newblock Survey of hallucination in natural language generation.
\newblock \emph{ACM Computing Surveys}.

\bibitem[{Jiang et~al.(2023)Jiang, Xu, Gao, Sun, Liu, Dwivedi-Yu, Yang, Callan,
  and Neubig}]{jiang2023active}
Zhengbao Jiang, Frank~F Xu, Luyu Gao, Zhiqing Sun, Qian Liu, Jane Dwivedi-Yu,
  Yiming Yang, Jamie Callan, and Graham Neubig. 2023.
\newblock Active retrieval augmented generation.
\newblock \emph{arXiv preprint arXiv:2305.06983}.

\bibitem[{Joshi et~al.(2017)Joshi, Choi, Weld, and Zettlemoyer}]{triviaqa}
Mandar Joshi, Eunsol Choi, Daniel~S. Weld, and Luke Zettlemoyer. 2017.
\newblock Triviaqa: {A} large scale distantly supervised challenge dataset for
  reading comprehension.
\newblock In \emph{Proceedings of the 55th Annual Meeting of the Association
  for Computational Linguistics, {ACL} 2017, Vancouver, Canada, July 30 -
  August 4, Volume 1: Long Papers}, pages 1601--1611. Association for
  Computational Linguistics.

\bibitem[{Kadavath et~al.(2022)Kadavath, Conerly, Askell, Henighan, Drain,
  Perez, Schiefer, Dodds, DasSarma, Tran-Johnson et~al.}]{kadavath2022language}
Saurav Kadavath, Tom Conerly, Amanda Askell, Tom Henighan, Dawn Drain, Ethan
  Perez, Nicholas Schiefer, Zac~Hatfield Dodds, Nova DasSarma, Eli
  Tran-Johnson, et~al. 2022.
\newblock Language models (mostly) know what they know.
\newblock \emph{arXiv preprint arXiv:2207.05221}.

\bibitem[{Karpukhin et~al.(2020)Karpukhin, O{\u{g}}uz, Min, Lewis, Wu, Edunov,
  Chen, and Yih}]{dpr}
Vladimir Karpukhin, Barlas O{\u{g}}uz, Sewon Min, Patrick Lewis, Ledell Wu,
  Sergey Edunov, Danqi Chen, and Wen-tau Yih. 2020.
\newblock Dense passage retrieval for open-domain question answering.
\newblock \emph{arXiv preprint arXiv:2004.04906}.

\bibitem[{Kasai et~al.(2022)Kasai, Sakaguchi, Takahashi, Bras, Asai, Yu, Radev,
  Smith, Choi, and Inui}]{realtimeQA}
Jungo Kasai, Keisuke Sakaguchi, Yoichi Takahashi, Ronan~Le Bras, Akari Asai,
  Xinyan Yu, Dragomir Radev, Noah~A Smith, Yejin Choi, and Kentaro Inui. 2022.
\newblock Realtime qa: What's the answer right now?
\newblock \emph{arXiv preprint arXiv:2207.13332}.

\bibitem[{Komeili et~al.(2021)Komeili, Shuster, and
  Weston}]{komeili2021internet}
Mojtaba Komeili, Kurt Shuster, and Jason Weston. 2021.
\newblock Internet-augmented dialogue generation.
\newblock \emph{arXiv preprint arXiv:2107.07566}.

\bibitem[{Kwiatkowski et~al.(2019)Kwiatkowski, Palomaki, Redfield, Collins,
  Parikh, Alberti, Epstein, Polosukhin, Devlin, Lee, Toutanova, Jones, Kelcey,
  Chang, Dai, Uszkoreit, Le, and Petrov}]{nq}
Tom Kwiatkowski, Jennimaria Palomaki, Olivia Redfield, Michael Collins,
  Ankur~P. Parikh, Chris Alberti, Danielle Epstein, Illia Polosukhin, Jacob
  Devlin, Kenton Lee, Kristina Toutanova, Llion Jones, Matthew Kelcey,
  Ming{-}Wei Chang, Andrew~M. Dai, Jakob Uszkoreit, Quoc Le, and Slav Petrov.
  2019.
\newblock Natural questions: a benchmark for question answering research.
\newblock \emph{Trans. Assoc. Comput. Linguistics}, 7:452--466.

\bibitem[{Lee et~al.(2019)Lee, Chang, and Toutanova}]{orqa}
Kenton Lee, Ming-Wei Chang, and Kristina Toutanova. 2019.
\newblock Latent retrieval for weakly supervised open domain question
  answering.
\newblock \emph{arXiv preprint arXiv:1906.00300}.

\bibitem[{Levy et~al.(2017)Levy, Seo, Choi, and Zettlemoyer}]{zsre}
Omer Levy, Minjoon Seo, Eunsol Choi, and Luke Zettlemoyer. 2017.
\newblock Zero-shot relation extraction via reading comprehension.
\newblock In \emph{Proceedings of the 21st Conference on Computational Natural
  Language Learning (CoNLL 2017), Vancouver, Canada, August 3-4, 2017}, pages
  333--342. Association for Computational Linguistics.

\bibitem[{Lewis et~al.(2020{\natexlab{a}})Lewis, Liu, Goyal, Ghazvininejad,
  Mohamed, Levy, Stoyanov, and Zettlemoyer}]{bart}
Mike Lewis, Yinhan Liu, Naman Goyal, Marjan Ghazvininejad, Abdelrahman Mohamed,
  Omer Levy, Veselin Stoyanov, and Luke Zettlemoyer. 2020{\natexlab{a}}.
\newblock \href {https://doi.org/10.18653/v1/2020.acl-main.703} {{BART:}
  denoising sequence-to-sequence pre-training for natural language generation,
  translation, and comprehension}.
\newblock In \emph{Proceedings of the 58th Annual Meeting of the Association
  for Computational Linguistics, {ACL} 2020, Online, July 5-10, 2020}, pages
  7871--7880. Association for Computational Linguistics.

\bibitem[{Lewis et~al.(2020{\natexlab{b}})Lewis, Perez, Piktus, Petroni,
  Karpukhin, Goyal, K{\"u}ttler, Lewis, Yih, Rockt{\"a}schel et~al.}]{rag}
Patrick Lewis, Ethan Perez, Aleksandra Piktus, Fabio Petroni, Vladimir
  Karpukhin, Naman Goyal, Heinrich K{\"u}ttler, Mike Lewis, Wen-tau Yih, Tim
  Rockt{\"a}schel, et~al. 2020{\natexlab{b}}.
\newblock Retrieval-augmented generation for knowledge-intensive nlp tasks.
\newblock \emph{Advances in Neural Information Processing Systems},
  33:9459--9474.

\bibitem[{Lin et~al.(2020)Lin, Lee, Khanna, and Ren}]{numersense}
Bill~Yuchen Lin, Seyeon Lee, Rahul Khanna, and Xiang Ren. 2020.
\newblock Birds have four legs?! numersense: Probing numerical commonsense
  knowledge of pre-trained language models.
\newblock In \emph{Proceedings of EMNLP}.
\newblock To appear.

\bibitem[{Lin(2004)}]{lin2004rouge}
Chin-Yew Lin. 2004.
\newblock Rouge: A package for automatic evaluation of summaries.
\newblock In \emph{Text summarization branches out}, pages 74--81.

\bibitem[{Liu et~al.(2019{\natexlab{a}})Liu, Ott, Goyal, Du, Joshi, Chen, Levy,
  Lewis, Zettlemoyer, and Stoyanov}]{roberta}
Yinhan Liu, Myle Ott, Naman Goyal, Jingfei Du, Mandar Joshi, Danqi Chen, Omer
  Levy, Mike Lewis, Luke Zettlemoyer, and Veselin Stoyanov. 2019{\natexlab{a}}.
\newblock \href {http://arxiv.org/abs/1910.10683} {Roberta: A robustly
  optimized bert pretraining approach}.
\newblock \emph{arXiv preprint arXiv:1907.11692}.

\bibitem[{Liu et~al.(2019{\natexlab{b}})Liu, Xiong, Sun, and Liu}]{liu2019fine}
Zhenghao Liu, Chenyan Xiong, Maosong Sun, and Zhiyuan Liu. 2019{\natexlab{b}}.
\newblock Fine-grained fact verification with kernel graph attention network.
\newblock \emph{arXiv preprint arXiv:1910.09796}.

\bibitem[{Loshchilov and Hutter(2019)}]{adamw}
Ilya Loshchilov and Frank Hutter. 2019.
\newblock \href {https://openreview.net/forum?id=Bkg6RiCqY7} {Decoupled weight
  decay regularization}.
\newblock In \emph{International Conference on Learning Representations}.

\bibitem[{Luccioni and Viviano(2021)}]{luccioni2021s}
Alexandra Luccioni and Joseph Viviano. 2021.
\newblock What’s in the box? an analysis of undesirable content in the common
  crawl corpus.
\newblock In \emph{Proceedings of the 59th Annual Meeting of the Association
  for Computational Linguistics and the 11th International Joint Conference on
  Natural Language Processing (Volume 2: Short Papers)}, pages 182--189.

\bibitem[{Maillard et~al.(2021)Maillard, Karpukhin, Petroni, Yih, O{\u{g}}uz,
  Stoyanov, and Ghosh}]{maillard2021multi}
Jean Maillard, Vladimir Karpukhin, Fabio Petroni, Wen-tau Yih, Barlas
  O{\u{g}}uz, Veselin Stoyanov, and Gargi Ghosh. 2021.
\newblock Multi-task retrieval for knowledge-intensive tasks.
\newblock \emph{arXiv preprint arXiv:2101.00117}.

\bibitem[{Menick et~al.(2022)Menick, Trebacz, Mikulik, Aslanides, Song,
  Chadwick, Glaese, Young, Campbell-Gillingham, Irving
  et~al.}]{menick2022teaching}
Jacob Menick, Maja Trebacz, Vladimir Mikulik, John Aslanides, Francis Song,
  Martin Chadwick, Mia Glaese, Susannah Young, Lucy Campbell-Gillingham,
  Geoffrey Irving, et~al. 2022.
\newblock Teaching language models to support answers with verified quotes.
\newblock \emph{arXiv preprint arXiv:2203.11147}.

\bibitem[{Nakano et~al.(2021)Nakano, Hilton, Balaji, Wu, Ouyang, Kim, Hesse,
  Jain, Kosaraju, Saunders et~al.}]{webgpt}
Reiichiro Nakano, Jacob Hilton, Suchir Balaji, Jeff Wu, Long Ouyang, Christina
  Kim, Christopher Hesse, Shantanu Jain, Vineet Kosaraju, William Saunders,
  et~al. 2021.
\newblock Webgpt: Browser-assisted question-answering with human feedback.
\newblock \emph{arXiv preprint arXiv:2112.09332}.

\bibitem[{Petroni et~al.(2020)Petroni, Piktus, Fan, Lewis, Yazdani, De~Cao,
  Thorne, Jernite, Karpukhin, Maillard et~al.}]{kilt}
Fabio Petroni, Aleksandra Piktus, Angela Fan, Patrick Lewis, Majid Yazdani,
  Nicola De~Cao, James Thorne, Yacine Jernite, Vladimir Karpukhin, Jean
  Maillard, et~al. 2020.
\newblock Kilt: a benchmark for knowledge intensive language tasks.
\newblock \emph{arXiv preprint arXiv:2009.02252}.

\bibitem[{Petroni et~al.(2019)Petroni, Rockt{\"a}schel, Lewis, Bakhtin, Wu,
  Miller, and Riedel}]{petroni2019language}
Fabio Petroni, Tim Rockt{\"a}schel, Patrick Lewis, Anton Bakhtin, Yuxiang Wu,
  Alexander~H Miller, and Sebastian Riedel. 2019.
\newblock Language models as knowledge bases?
\newblock \emph{arXiv preprint arXiv:1909.01066}.

\bibitem[{Piktus et~al.(2021)Piktus, Petroni, Karpukhin, Okhonko, Broscheit,
  Izacard, Lewis, O{\u{g}}uz, Grave, Yih et~al.}]{sphere}
Aleksandra Piktus, Fabio Petroni, Vladimir Karpukhin, Dmytro Okhonko, Samuel
  Broscheit, Gautier Izacard, Patrick Lewis, Barlas O{\u{g}}uz, Edouard Grave,
  Wen-tau Yih, et~al. 2021.
\newblock The web is your oyster--knowledge-intensive nlp against a very large
  web corpus.
\newblock \emph{arXiv preprint arXiv:2112.09924}.

\bibitem[{Raffel et~al.(2020)Raffel, Shazeer, Roberts, Lee, Narang, Matena,
  Zhou, Li, and Liu}]{t5}
Colin Raffel, Noam Shazeer, Adam Roberts, Katherine Lee, Sharan Narang, Michael
  Matena, Yanqi Zhou, Wei Li, and Peter~J. Liu. 2020.
\newblock \href {http://jmlr.org/papers/v21/20-074.html} {Exploring the limits
  of transfer learning with a unified text-to-text transformer}.
\newblock \emph{J. Mach. Learn. Res.}, 21:140:1--140:67.

\bibitem[{Rajpurkar et~al.(2016)Rajpurkar, Zhang, Lopyrev, and Liang}]{squad}
Pranav Rajpurkar, Jian Zhang, Konstantin Lopyrev, and Percy Liang. 2016.
\newblock \href {https://doi.org/10.18653/v1/d16-1264} {Squad: 100, 000+
  questions for machine comprehension of text}.
\newblock In \emph{Proceedings of the 2016 Conference on Empirical Methods in
  Natural Language Processing, {EMNLP} 2016, Austin, Texas, USA, November 1-4,
  2016}, pages 2383--2392. The Association for Computational Linguistics.

\bibitem[{Ren et~al.(2021)Ren, Xiao, Chang, Huang, Li, Gupta, Chen, and
  Wang}]{ren2021survey}
Pengzhen Ren, Yun Xiao, Xiaojun Chang, Po-Yao Huang, Zhihui Li, Brij~B Gupta,
  Xiaojiang Chen, and Xin Wang. 2021.
\newblock A survey of deep active learning.
\newblock \emph{ACM computing surveys (CSUR)}, 54(9):1--40.

\bibitem[{Roberts et~al.(2020)Roberts, Raffel, and Shazeer}]{roberts2020much}
Adam Roberts, Colin Raffel, and Noam Shazeer. 2020.
\newblock How much knowledge can you pack into the parameters of a language
  model?
\newblock \emph{arXiv preprint arXiv:2002.08910}.

\bibitem[{Robertson et~al.(2009)Robertson, Zaragoza et~al.}]{bm25}
Stephen Robertson, Hugo Zaragoza, et~al. 2009.
\newblock The probabilistic relevance framework: Bm25 and beyond.
\newblock \emph{Foundations and Trends{\textregistered} in Information
  Retrieval}, 3(4):333--389.

\bibitem[{Sakaguchi et~al.(2020)Sakaguchi, Bras, Bhagavatula, and
  Choi}]{winogrande}
Keisuke Sakaguchi, Ronan~Le Bras, Chandra Bhagavatula, and Yejin Choi. 2020.
\newblock Winogrande: An adversarial winograd schema challenge at scale.
\newblock In \emph{The Thirty-Fourth {AAAI} Conference on Artificial
  Intelligence, {AAAI} 2020, The Thirty-Second Innovative Applications of
  Artificial Intelligence Conference, {IAAI} 2020, The Tenth {AAAI} Symposium
  on Educational Advances in Artificial Intelligence, {EAAI} 2020, New York,
  NY, USA, February 7-12, 2020}, pages 8732--8740. {AAAI} Press.

\bibitem[{Sang and De~Meulder(2003)}]{sang2003introduction}
Erik~F Sang and Fien De~Meulder. 2003.
\newblock Introduction to the conll-2003 shared task: Language-independent
  named entity recognition.
\newblock \emph{arXiv preprint cs/0306050}.

\bibitem[{Sap et~al.(2019)Sap, Rashkin, Chen, Bras, and Choi}]{socialiqa}
Maarten Sap, Hannah Rashkin, Derek Chen, Ronan~Le Bras, and Yejin Choi. 2019.
\newblock Social iqa: Commonsense reasoning about social interactions.
\newblock In \emph{Proceedings of the 2019 Conference on Empirical Methods in
  Natural Language Processing and the 9th International Joint Conference on
  Natural Language Processing, {EMNLP-IJCNLP} 2019, Hong Kong, China, November
  3-7, 2019}, pages 4462--4472. Association for Computational Linguistics.

\bibitem[{Shuster et~al.(2020)Shuster, Ju, Roller, Dinan, Boureau, and
  Weston}]{eli5}
Kurt Shuster, Da~Ju, Stephen Roller, Emily Dinan, Y{-}Lan Boureau, and Jason
  Weston. 2020.
\newblock The dialogue dodecathlon: Open-domain knowledge and image grounded
  conversational agents.
\newblock In \emph{Proceedings of the 58th Annual Meeting of the Association
  for Computational Linguistics, {ACL} 2020, Online, July 5-10, 2020}, pages
  2453--2470. Association for Computational Linguistics.

\bibitem[{Storks et~al.(2019)Storks, Gao, and Chai}]{storks2019recent}
Shane Storks, Qiaozi Gao, and Joyce~Y Chai. 2019.
\newblock Recent advances in natural language inference: A survey of
  benchmarks, resources, and approaches.
\newblock \emph{arXiv preprint arXiv:1904.01172}.

\bibitem[{Surdeanu and Ji(2014)}]{surdeanu2014overview}
Mihai Surdeanu and Heng Ji. 2014.
\newblock Overview of the english slot filling track at the tac2014 knowledge
  base population evaluation.
\newblock In \emph{Proc. Text Analysis Conference (TAC2014)}.

\bibitem[{Talmor et~al.(2019)Talmor, Herzig, Lourie, and Berant}]{csqa}
Alon Talmor, Jonathan Herzig, Nicholas Lourie, and Jonathan Berant. 2019.
\newblock Commonsenseqa: {A} question answering challenge targeting commonsense
  knowledge.
\newblock In \emph{Proceedings of the 2019 Conference of the North American
  Chapter of the Association for Computational Linguistics: Human Language
  Technologies, {NAACL-HLT} 2019, Minneapolis, MN, USA, June 2-7, 2019, Volume
  1 (Long and Short Papers)}, pages 4149--4158. Association for Computational
  Linguistics.

\bibitem[{Tang et~al.(2022)Tang, Li, Zhao, and Wen}]{mvp}
Tianyi Tang, Junyi Li, Wayne~Xin Zhao, and Ji-Rong Wen. 2022.
\newblock Mvp: Multi-task supervised pre-training for natural language
  generation.
\newblock \emph{arXiv preprint arXiv:2206.12131}.

\bibitem[{Thoppilan et~al.(2022)Thoppilan, De~Freitas, Hall, Shazeer,
  Kulshreshtha, Cheng, Jin, Bos, Baker, Du et~al.}]{thoppilan2022lamda}
Romal Thoppilan, Daniel De~Freitas, Jamie Hall, Noam Shazeer, Apoorv
  Kulshreshtha, Heng-Tze Cheng, Alicia Jin, Taylor Bos, Leslie Baker, Yu~Du,
  et~al. 2022.
\newblock Lamda: Language models for dialog applications.
\newblock \emph{arXiv preprint arXiv:2201.08239}.

\bibitem[{Thorne et~al.(2018)Thorne, Vlachos, Christodoulopoulos, and
  Mittal}]{fever}
James Thorne, Andreas Vlachos, Christos Christodoulopoulos, and Arpit Mittal.
  2018.
\newblock {FEVER}: a large-scale dataset for fact extraction and
  {VERification}.
\newblock In \emph{NAACL-HLT}.

\bibitem[{Wagner et~al.(2016)Wagner, Graells-Garrido, Garcia, and
  Menczer}]{wagner2016women}
Claudia Wagner, Eduardo Graells-Garrido, David Garcia, and Filippo Menczer.
  2016.
\newblock Women through the glass ceiling: gender asymmetries in wikipedia.
\newblock \emph{EPJ Data Science}, 5:1--24.

\bibitem[{Wenzek et~al.(2020)Wenzek, Lachaux, Conneau, Chaudhary, Guzm{\'{a}}n,
  Joulin, and Grave}]{ccnet}
Guillaume Wenzek, Marie{-}Anne Lachaux, Alexis Conneau, Vishrav Chaudhary,
  Francisco Guzm{\'{a}}n, Armand Joulin, and Edouard Grave. 2020.
\newblock Ccnet: Extracting high quality monolingual datasets from web crawl
  data.
\newblock In \emph{Proceedings of The 12th Language Resources and Evaluation
  Conference, {LREC} 2020, Marseille, France, May 11-16, 2020}, pages
  4003--4012. European Language Resources Association.

\bibitem[{Wu et~al.(2019)Wu, Petroni, Josifoski, Riedel, and
  Zettlemoyer}]{wu2019scalable}
Ledell Wu, Fabio Petroni, Martin Josifoski, Sebastian Riedel, and Luke
  Zettlemoyer. 2019.
\newblock Scalable zero-shot entity linking with dense entity retrieval.
\newblock \emph{arXiv preprint arXiv:1911.03814}.

\bibitem[{Yang et~al.(2018)Yang, Qi, Zhang, Bengio, Cohen, Salakhutdinov, and
  Manning}]{hotpotqa}
Zhilin Yang, Peng Qi, Saizheng Zhang, Yoshua Bengio, William~W. Cohen, Ruslan
  Salakhutdinov, and Christopher~D. Manning. 2018.
\newblock {HotpotQA}: A dataset for diverse, explainable multi-hop question
  answering.
\newblock In \emph{Conference on Empirical Methods in Natural Language
  Processing ({EMNLP})}.

\bibitem[{Yin et~al.(2022)Yin, Dong, Cheng, Liu, Chang, Wei, and
  Gao}]{kit_survey}
Da~Yin, Li~Dong, Hao Cheng, Xiaodong Liu, Kai-Wei Chang, Furu Wei, and Jianfeng
  Gao. 2022.
\newblock A survey of knowledge-intensive nlp with pre-trained language models.
\newblock \emph{arXiv preprint arXiv:2202.08772}.

\bibitem[{Zellers et~al.(2019)Zellers, Holtzman, Bisk, Farhadi, and
  Choi}]{hellaswag}
Rowan Zellers, Ari Holtzman, Yonatan Bisk, Ali Farhadi, and Yejin Choi. 2019.
\newblock Hellaswag: Can a machine really finish your sentence?
\newblock In \emph{Proceedings of the 57th Conference of the Association for
  Computational Linguistics, {ACL} 2019, Florence, Italy, July 28- August 2,
  2019, Volume 1: Long Papers}, pages 4791--4800. Association for Computational
  Linguistics.

\bibitem[{Zhang et~al.(2020)Zhang, Sun, Galley, Chen, Brockett, Gao, Gao, Liu,
  and Dolan}]{dialogpt}
Yizhe Zhang, Siqi Sun, Michel Galley, Yen-Chun Chen, Chris Brockett, Xiang Gao,
  Jianfeng Gao, Jingjing Liu, and Bill Dolan. 2020.
\newblock \href {https://doi.org/10.18653/v1/2020.acl-demos.30} {{DIALOGPT} :
  Large-scale generative pre-training for conversational response generation}.
\newblock In \emph{Proceedings of the 58th Annual Meeting of the Association
  for Computational Linguistics: System Demonstrations}, pages 270--278,
  Online. Association for Computational Linguistics.

\bibitem[{Zhao et~al.(2022)Zhao, Liu, Ren, and Wen}]{dense-survey}
Wayne~Xin Zhao, Jing Liu, Ruiyang Ren, and Ji{-}Rong Wen. 2022.
\newblock \href {https://doi.org/10.48550/arXiv.2211.14876} {Dense text
  retrieval based on pretrained language models: {A} survey}.
\newblock \emph{CoRR}, abs/2211.14876.

\bibitem[{Zhao et~al.(2023)Zhao, Zhou, Li, Tang, Wang, Hou, Min, Zhang, Zhang,
  Dong, Du, Yang, Chen, Chen, Jiang, Ren, Li, Tang, Liu, Liu, Nie, and
  Wen}]{LLM-survey}
Wayne~Xin Zhao, Kun Zhou, Junyi Li, Tianyi Tang, Xiaolei Wang, Yupeng Hou,
  Yingqian Min, Beichen Zhang, Junjie Zhang, Zican Dong, Yifan Du, Chen Yang,
  Yushuo Chen, Zhipeng Chen, Jinhao Jiang, Ruiyang Ren, Yifan Li, Xinyu Tang,
  Zikang Liu, Peiyu Liu, Jian{-}Yun Nie, and Ji{-}Rong Wen. 2023.
\newblock \href {https://doi.org/10.48550/arXiv.2303.18223} {A survey of large
  language models}.
\newblock \emph{CoRR}, abs/2303.18223.

\bibitem[{Zhu et~al.(2021)Zhu, Lei, Wang, Zheng, Poria, and
  Chua}]{zhu2021retrieving}
Fengbin Zhu, Wenqiang Lei, Chao Wang, Jianming Zheng, Soujanya Poria, and
  Tat-Seng Chua. 2021.
\newblock Retrieving and reading: A comprehensive survey on open-domain
  question answering.
\newblock \emph{arXiv preprint arXiv:2101.00774}.

\end{thebibliography}
\bibliographystyle{acl_natbib}

\newpage
\appendix

\section*{Appendix}
We provide some experiment-related information as supplementary materials. The appendix is organized into three sections:
\begin{itemize}
	\item Details of pretraining tasks are presented in Appendix~\ref{app-dataset};
	\item Model architecture and pretraining details are presented in Appendix~\ref{app-configuration};
        \item Supplementary experiments are presented in Appendix~\ref{app-experiment};
	\item Examples with retrieved knowledge are presented in Appendix~\ref{app-examples}.
\end{itemize}

\section{Pretraining Tasks} \label{app-dataset}

As described in Section~\ref{sec-unification}, to pretrain our model, we unify 16 knowledge-intensive tasks across seven categories into a general text-to-text format:

\begin{itemize}[leftmargin=*,itemsep=-1.5pt,topsep=4pt]
    \item \textbf{Fact checking} is the task of assessing whether a natural language claim is true~\cite{guo2022survey}. It requires deep knowledge about the claim. We consider the claim as \textit{input} and the classification label (\eg true/false) as \textit{output}.
    \item \textbf{Slot filling} aims to complete the missing information for certain relations of entities~\cite{surdeanu2014overview} (\eg subject entity \textit{Star Trek} and relation \textit{creator}). It requires entity disambiguation and the relational knowledge for entities. We model the structured string ``subject entity [SEP] relation'' as \textit{input} and the object entity as \textit{output}.
    \item \textbf{Dialogue} focuses on building an engaging chatbot that can discusses a wide range of open-ended topics such as whether~\cite{huang2020challenges}. It requires models to know about the background knowledge for the conversational topics. We consider the dialogue history as \textit{input} and the next utterance as \textit{output}.
    \item \textbf{Open-domain question answering} is the task of producing answers to factoid questions in natural language~\cite{zhu2021retrieving}. The questions could be about nearly anything relying on world knowledge. We consider the question as \textit{input} and the answer as \textit{output}.
    \item \textbf{Commonsense question answering} aims to test if models can answer questions regarding commonsense knowledge that everyone knows~\cite{dou2022zero}. Similarly, we consider the question as \textit{input} and the answer as \textit{output}.
    \item \textbf{Commonsense reasoning} is intended to utilize commonsense knowledge to reason about certain aspects of the given text~\cite{winogrande}. Therefore, we consider the given text as \textit{input} and the prediction as \textit{output}.
    \item \textbf{Natural language inference} is the task of determining whether the given ``hypothesis'' logically follows from the ``premise''~\cite{storks2019recent}. It acquires deep knowledge about the relationship between hypothesis and premise. We consider the premise as \textit{input} and the hypothesis as \textit{output}.
\end{itemize}

For each category, we choose several representative tasks to construct our pretraining corpus. The detailed information of these included tasks is listed in Table~\ref{tab-data}. To mitigate the huge disparity between dataset sizes, we follow \cite{t5} to use the temperature-scaled mixing strategy with a rate of $T=2$ for setting the proportion of data coming from each task.  During pretraining, for each task example, we use BM25 to retrieve top-$10$ passages from CCNet as our external knowledge. The input texts are concatenated with the retrieved passages using manually-written prompts. The final input is constructed in the following format:
\begin{align}
    &\text{\textbf{Context:}~[passage$_1$]...[passage$_{10}$]} \nonumber \\
    &\text{[Task Instruction]:~[the original input text]} \nonumber \\
    &\text{\textbf{Option $1$:} [option$_1$]...\textbf{Option $n$:} [option$_n$]} \nonumber
\end{align}
The ``Option'' string is applied only when the input text is provided with several candidate answers. The blanks ``[passage$_n$]'' and ``[option$_n$]'' is filled with the retrieved passages and candidate answers. The blank ``[Task Instruction]'' aims to indicate the task for our model, which is task-specific and detailed in Table~\ref{tab-instruction}.

\begin{table*}[t]
	\centering
	\small
	\begin{tabular}{c l r r r}
		\toprule[1pt]
		\textbf{Task Families} & \textbf{Tasks} & \textbf{\#Train} & \textbf{\#Validation}  & \textbf{\#Test}\\
            \midrule[0.5pt]
            \tabincell{c}{Fact Checking} & FEVER~\cite{fever} & 134,287 & 14,342 & 10,100 \\
		\midrule[0.5pt]
		\multirow{2.5}*{\tabincell{c}{Slot Filling}} & T-REx~\cite{trex} & 2,999,272 & 26,833 & 5,000 \\
            \cmidrule[0.5pt]{2-5}
            & zsRE~\cite{zsre} & 154,826 & 3,771 & 4,966 \\
            \midrule[0.5pt]
            \tabincell{c}{Dialogue} & WoW~\cite{wow} & 63,734 & 3,054 & 2,944 \\
            \midrule[0.5pt]
            \multirow{5.5}*{\tabincell{c}{Open-domain QA}} & NQ~\cite{nq} & 108,890 & 6,008 & 1,444 \\
            \cmidrule[0.5pt]{2-5}
            & TriviaQA~\cite{triviaqa} & 1,835,943 & 168,358 & 6,586 \\
            \cmidrule[0.5pt]{2-5}
            & HotpotQA~\cite{hotpotqa} & 88,869 & 5,600 & 5,569 \\
            \cmidrule[0.5pt]{2-5}
            & ELI5~\cite{eli5} & 804,370 & 18,037 & 600 \\
            \midrule[0.5pt]
            \multirow{5.5}*{\tabincell{c}{Commonsense QA}} & CSQA~\cite{csqa} & 9,741 & 1,221 & 1,140 \\
            \cmidrule[0.5pt]{2-5}
            & SocialIQa~\cite{socialiqa} & 33,410 & 1,954 & 2,059 \\
            \cmidrule[0.5pt]{2-5}
            & CosmosQA~\cite{cosmosqa} & 25,262 & 2,985 & 6,963 \\
            \cmidrule[0.5pt]{2-5}
            & PIQA~\cite{piqa} & 16,113 & 1,838 & 3,084 \\
            \midrule[0.5pt]
            \multirow{2.5}*{\tabincell{c}{Commonsense Reasoning}} & NumerSense~\cite{numersense} & 10,444 & 200 & 3,146 \\
            \cmidrule[0.5pt]{2-5}
            & WinoGrande~\cite{winogrande} & 40,398 & 1,267 & 1,767 \\
            \midrule[0.5pt]
            \multirow{2.5}*{\tabincell{c}{Natural Language Inference}}  & HellaSwag~\cite{hellaswag} & 39,905 & 10,042 & 10,003 \\
            \cmidrule[0.5pt]{2-5}
            & $\alpha$NLI~\cite{anli} & 169,654 & 1,532 & 3,059 \\
		\bottomrule[1pt]
	\end{tabular}%
	\caption{The statistics of our 16 knowledge-intensive tasks.}
	\label{tab-data}%
\end{table*}

\begin{table}[h]
\renewcommand\arraystretch{1.1}
\setlength\tabcolsep{3pt}
	\centering
	\small
	\begin{tabular}{c|l}
		\toprule[1pt]
		\textbf{Tasks} & \textbf{Task Instructions} \\
		\midrule[0.7pt]
		Fact Checking & Verify the following claim  \\
            \midrule[0.5pt]
		Slot Filling & Predict the missing fact  \\
            \midrule[0.5pt]
		\tabincell{c}{Open-domain\\QA} & Answer the following question  \\
            \midrule[0.5pt]
		\tabincell{c}{Commonsense\\QA} & Answer the following question  \\
        \midrule[0.5pt]
            Dialogue & Response to the following dialogue \\
            \midrule[0.5pt]
            \tabincell{c}{Natural Language\\Inference} & Inference on the following context \\
            \midrule[0.5pt]
            \tabincell{c}{Commonsense\\Reasoning} & Reason about the following sentence \\
		\bottomrule[1pt]
	\end{tabular}%
	\caption{Task instructions for each task category.}
	\label{tab-instruction}%
\end{table}

\section{Implementation Details} \label{app-configuration}

Our UniWeb model uses a Transformer with $12$ layers in both encoder and decoder ($406$M parameters), the same as the model size of BART\textsubscript{\textsc{large}}~\citep{bart}. The hidden size is $1{,}024$ and the inner hidden size of the feed-forward network is $4{,}096$. We employ the byte-pair-encoding (BPE) tokenizer, and the vocabulary size is $50{,}267$. We initialize the backbone with the MVP model~\cite{mvp}, a supervised pre-trained PLM, to provide a good starting point for generation following previous work~\citep{unilm,dialogpt}. We pre-train the model with batch size $8{,}192$ on Tesla A$100$ $40$GB GPUs. 

For our model, the maximum length of both input and output sequences is set to $1{,}024$ for supporting examples to contain more tokens. We optimize the model with a constant learning rate of $2\times10^{-5}$ using standard sequence-to-sequence cross-entropy loss. We apply the AdamW optimizer~\citep{adamw} with $\beta_1=0.9$, $\beta_2=0.98$, $\epsilon=1 \times 10^{-6}$ to improve training stability~\citep{roberta}. The weight decay coefficient is $0.1$. For testing, we select the checkpoint with the highest validation performance. 
According to the results shown in Figure~\ref{fig:sensitivity}(a), we set the entropy threshold $\eta$ as 4.0.

Since the tasks of fact checking, slot filling, dialogue, and open-domain QA are specially designed based on the knowledge from Wikipedia, we require the search engine to retrieve the top-1 passage from the website \url{https://en.wikipedia.org}.

\section{Supplementary Experiments}
\label{app-experiment}

\begin{table}[t]
	\small
	\centering
	\begin{tabular}{lcc}
		\toprule
		 \multirow{2.5}*{\textbf{Models}} & \multicolumn{2}{c}{\textbf{\textsc{RealTime QA}}} \\
		 \cmidrule{2-3}
		 & \textbf{Original} & \textbf{NOTA} \\
		\midrule
		T5       & 40.0          & 33.3       \\
            GPT-3 & 56.7 & 23.3 \\
            \cmidrule{1-3}
		RAG+DPR         & 10.0  & 16.7   \\
            RAG+Google Search & 63.3 & 50.0  \\
            \cmidrule{1-3}
		UniWeb +Google Search      & 66.7          & 56.7        \\
		\bottomrule
	\end{tabular}
	\caption{Accuracy results for the questions at week from 2022/12/11 through 2022/12/17. We utilize DPR to retrieve top-5 documents from Wikipedia and use Google Search to retrieve top-5 news articles.}
	\label{tab:real-time}
\end{table}

\paratitle{RealTime QA.} Previous QA systems mostly assume that answers are static regardless of the time of query~\cite{ChenY20}. In this section, we use the \textsc{RealTime QA} benchmark~\cite{realtimeQA} to test models about real-time, instantaneous information. At each week, \textsc{RealTime QA} will retrieve news articles and \textasciitilde30 human-written, multiple-choice questions from news websites (CNN, THE WEEK, and USA Today), which covers diverse topics such as politics, business, sports, and entertainment. We adopt the original and NOTA (none of the above) settings and test our models over questions from 2022/12/11 through 2022/12/17. The results are shown in Table~\ref{tab:real-time}. Since one of the original choices is randomly replaced with ``none of the above'', the NOTA setting results in a distinct performance degradation. Besides, due to the real-time nature of the questions, only using DPR to retrieve texts from static Wikipedia achieves worse results. Our UniWeb model with Google Search performs the best. This indicates that UniWeb can answer questions based on the real-time information, rather than relying on past information from pre-training.

\begin{figure}[t]
	\centering
	\subfigure[]{
		\centering
		\includegraphics[width=0.22\textwidth]{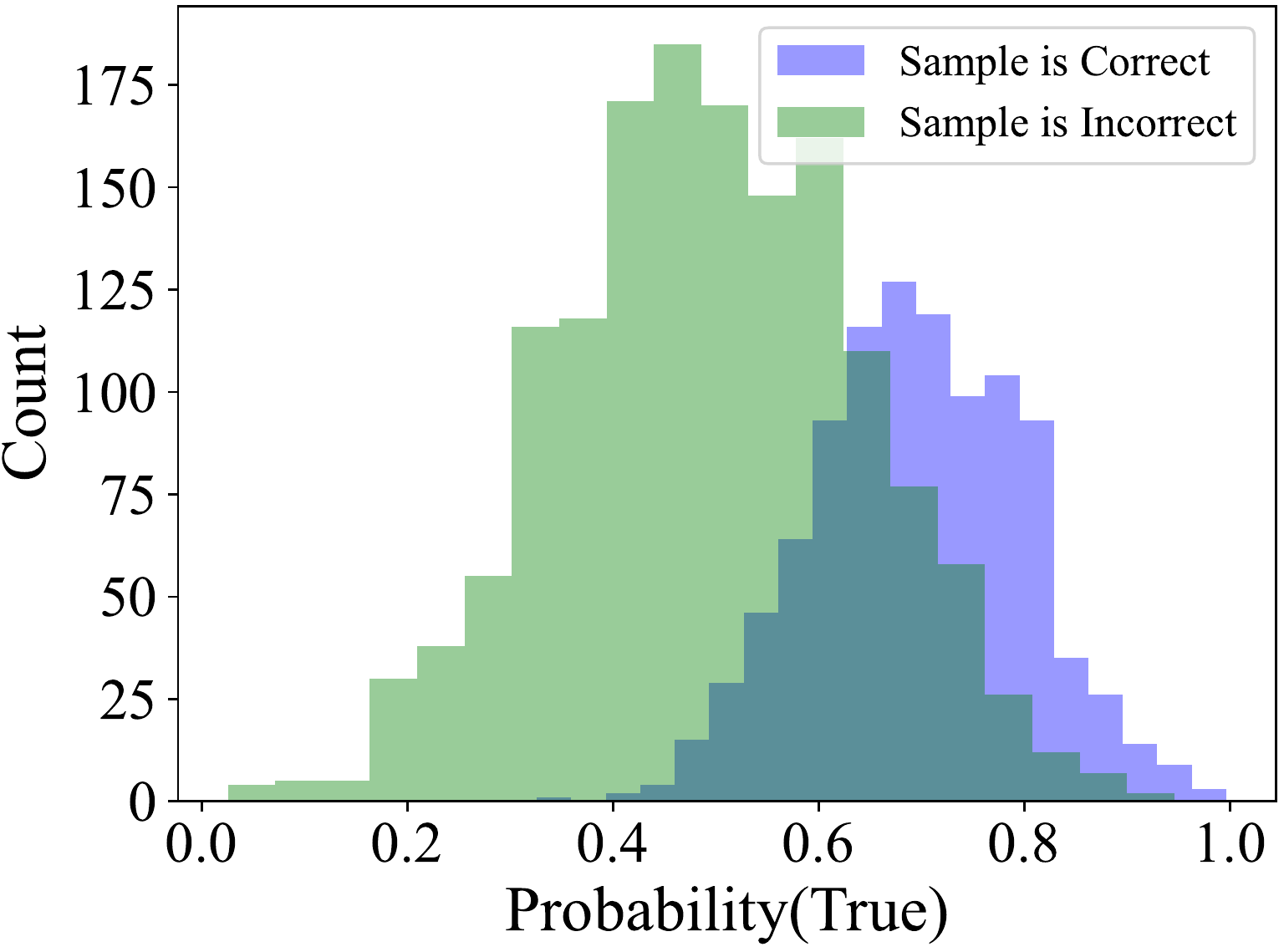}
	}
	\subfigure[]{
		\centering
		\includegraphics[width=0.22\textwidth]{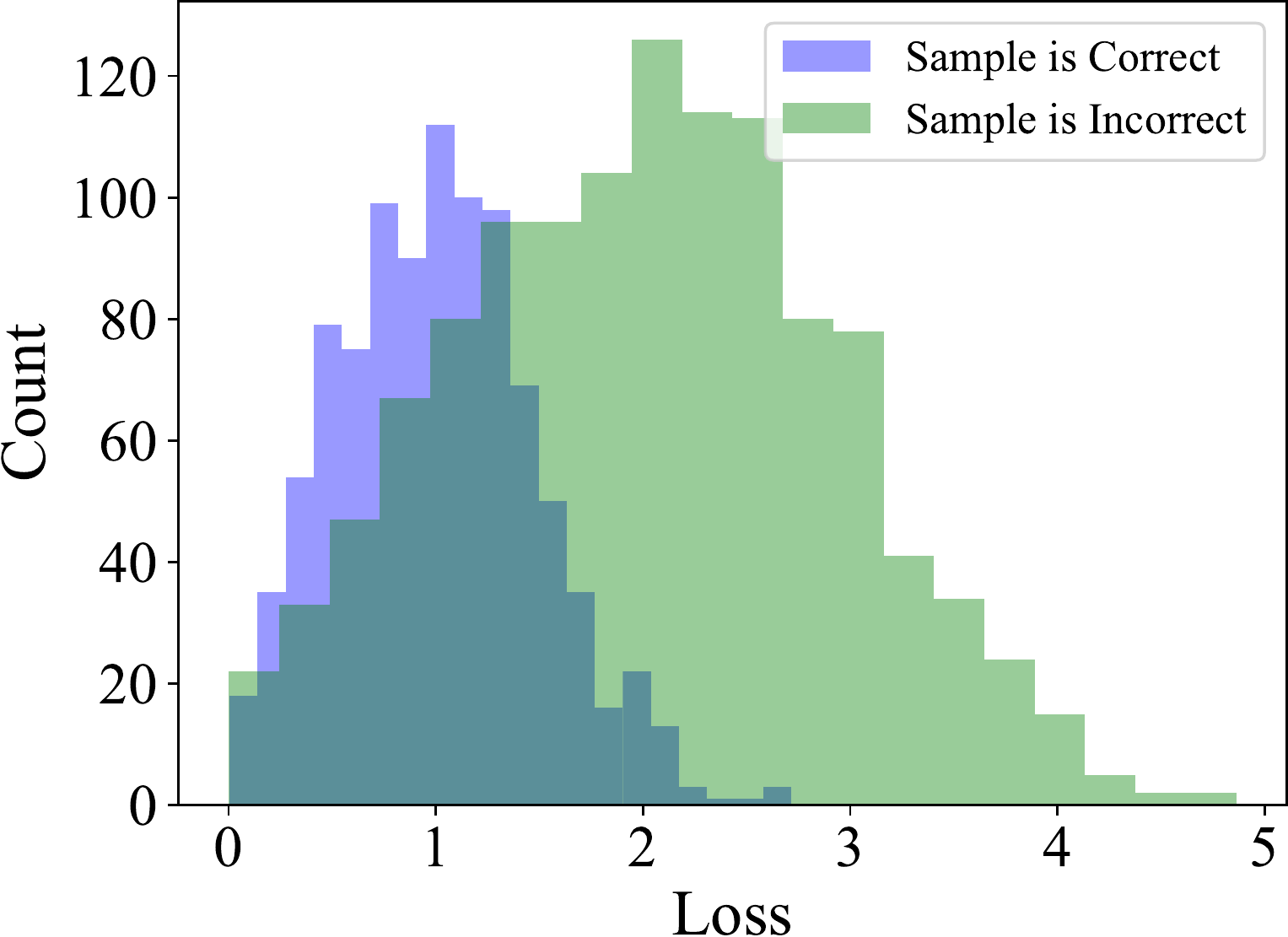}
	}
	\centering
	\caption{(a) Probability of True for prompts in HotpotQA; (b) Loss of samples in HotpotQA.}
	\label{fig:app-exp}
	\vspace{-0.2cm}
\end{figure}

\paratitle{Self-Evaluation Criteria.}
To evaluate the model confidence in task examples, we adopt the entropy as criterion in Section~\ref{sec-self-evaluation}. In this part, we test with more kinds of criteria compared to the entropy following \citet{kadavath2022language}. First, we consider a \emph{sample-enhanced prompting} method, where we generate five samples with beam search and ask the model about the validity of the first sample with the highest score. We show an example at below:
\begin{align}
    &\texttt{Question: Who is the third} \nonumber \\
    &\quad\texttt{president of the United States?} \nonumber \\
    &\texttt{Possible Answer: James Monroe} \nonumber \\
    &\texttt{Here are some brainstormed ideas:} \nonumber \\
    &\texttt{Thomas Jefferson} \nonumber \\
    &\texttt{John Adams} \nonumber \\
    &\texttt{Thomas Jefferson} \nonumber \\
    &\texttt{George Washington} \nonumber \\
    &\texttt{Is the possible answer:} \nonumber \\
    &~\texttt{(A) True} \nonumber \\
    &~\texttt{(B) False} \nonumber \\
    &\texttt{The possible answer is:} \nonumber
\end{align}
If the model self-evaluate the possible answer is \texttt{False}, our model will leverage the search engine to access the web, otherwise not. We show the probability of predicting \texttt{True} depending on whether the model gets the question correct in Figure~\ref{fig:app-exp}(a). However, according to \citet{kadavath2022language}, this self-evaluation method is mainly suitable for question answering tasks with short-form answers but benefits less on question answering tasks with long-form answers. Second, we consider using \emph{loss} as the criterion to evaluate the model confidence. This approach is to generate a sample, and then look at the model's loss on this sample, averaged over all tokens, like the knowledge-intensive learning loss (Eq.~\ref{eq-kil}). If the loss for an example is higher than a threshold (\eg 0.5), we consider that the model is unconfident about this example and we will query the web to retrieve knowledge. In Figure~\ref{fig:app-exp}(b), we show the loss of samples that the model gets correct or incorrect.

\section{Case Study} \label{app-examples}

In Table~\ref{tab:app-case-study}, we present three examples from TriviaQA~\cite{triviaqa}, CommonsenseQA~\cite{csqa}, and NumerSense~\cite{numersense}. The first TriviaQA dataset is specially designed based on the knowledge from Wikipedia. Therefore, we can observe that Wikipedia contains the most relevant passage about the topic ``US nuclear reactor accident in 1979''. In addition, the web can provide another source of knowledge about this topic. Although CCNet covers this content, it does not give a clear answer to this question (\ie full name of the US nuclear reactor). The second CommonsenseQA dataset involves questions related to commonsense knowledge going beyond Wikipedia. Therefore, Wikipedia can only provide a fuzzy description passage about ``Guitar''. The web and CCNet return diverse knowledge but the passage returned by search engine is more helpful. The thrid NumerSense dataset requires models to reason about the number. For the third example, CCNet provides a passage with incorrect information. While, the web and Wikipedia return passages about the rule of ``tic-tac-toe'', which can result in the correct answer ``three''. 

\begin{table*}[t]
	\small
	\centering
	\begin{tabular}{p{0.31\textwidth} p{0.31\textwidth} p{0.31\textwidth}}
		\toprule
		  \multicolumn{3}{l}{\tabincell{l}{\textbf{Question:} Which \textbf{US nuclear reactor} had a major \textbf{accident} in \textbf{1979}?}} \\
            \multicolumn{3}{l}{\textbf{Gold Answer:} Three Mile Island Unit 2 reactor} \\
		\midrule
            Top-1 Wikipedia Passage & Top-1 CCNet Passage & Top-1 Web Passage \\
            \midrule
            ...~The Three Mile Island \textcolor{red}{accident} was a partial meltdown of \textbf{the Three Mile Island, Unit 2 (TMI-2) reactor} in Pennsylvania, United States. It began at 4 a.m. on March 28, 1979. It is the most significant \textcolor{red}{accident in U.S.} commercial nuclear power plant history. On the seven-point International Nuclear Event Scale, it is rated Level 5 – Accident with Wider Consequences... \makecell[l]{ \url{https://en.wikipedia.org/wiki}\\\url{/Three_Mile_Island_accident}}
            &
            ...~The US and former Soviet Union had been operating nuclear power for 267 and 162 reactor-years respectively before a major accident occurred. At the time of \textcolor{red}{the Three Mile Island accident in 1979, the US} had 52 nuclear power stations, which had been operating for 267 reactor years, or an average of 5.1 years per reactor... \makecell[l]{ \url{https://chinadialogue.net/}\\\url{article/show/single/en/5808}\\\url{-Chinese-nuclear-disaster-}\\\url{highly-probable-by-2-3-}}
            &
            ...~\textbf{The Three Mile Island Unit 2 reactor}, near Middletown, Pa., partially melted down on March 28, 1979. This was the most serious \textcolor{red}{accident in U.S.} commercial nuclear power plant operating history, although its small radioactive releases had no detectable health effects on plant workers or the public... \makecell[l]{ \url{https://www.nrc.gov/reading}\\\url{-rm/doc-collections/fact-}\\\url{sheets/3mile-isle.html}} \\
            \toprule
		  \multicolumn{3}{l}{\tabincell{l}{\textbf{Question:} What do people typically do while \textbf{playing guitar}?}} \\
            \multicolumn{3}{l}{\textbf{Candidate Answers:} A. cry B. hear sounds C. singing D. arthritis E. making music} \\
            \multicolumn{3}{l}{\textbf{Gold Answer:} singing} \\
		\midrule
            Top-1 Wikipedia Passage & Top-1 CCNet Passage & Top-1 Web Passage \\
            \midrule
            ...~The \textcolor{red}{guitar} is a fretted musical instrument that typically has six strings. It is usually held flat against the player's body and played by strumming or plucking the strings with the dominant hand, while simultaneously pressing selected strings against frets with the fingers of the opposite hand. A plectrum or individual finger picks may also be used to strike the strings... \makecell[l]{ \url{https://en.wikipedia.org/}\\\url{wiki/Guitar}}
            &
            ...~I was playing a brand-new game that had no rules and nothing established. I was really shy about it at first, because I hadn't looked out into the world to find other people who, of course, had done things like this. I heard Fred Frith play, and I knew he played his \textcolor{red}{guitar} with objects not typically associated with the guitar... \makecell[l]{ \url{https://www.premierguitar.}\\\url{com/articles/24026-janet-}\\\url{feder-prepared-for-all-genres}}
            &
            ...~Practicing the \textcolor{red}{guitar} regularly can enhance your concentration and expand your attention span. It takes an adequate focus to become an expert guitarist. Focusing becomes a habit for your mind and will help you concentrate better on other everyday chores too... \makecell[l]{ \url{https://www.chasingsound.}\\\url{com/posts/10-health-benefits}\\\url{-of-playing-guitar}} \\
            \toprule
		  \multicolumn{3}{l}{\tabincell{l}{\textbf{Question:} How do you win at \textbf{tic-tac-toe} get <mask> of your symbols in a row?}} \\
            \multicolumn{3}{l}{\textbf{Gold Answer:} three} \\
		\midrule
            Top-1 Wikipedia Passage & Top-1 CCNet Passage & Top-1 Web Passage \\
            \midrule
            ...~Tic-tac-toe (American English), noughts and crosses (Commonwealth English), or Xs and Os (Canadian or Irish English) is a paper-and-pencil game for two players who take turns marking the spaces in a three-by-three grid with X or O. The player who succeeds in placing \textcolor{red}{three of their marks} in a horizontal, vertical, or diagonal row is the winner... \makecell[l]{ \url{https://en.wikipedia.org/}\\\url{wiki/Tic-tac-toe}}
            &
            ...~You just make a 4x4 box instead of a 3x3 box. Then the same rules apply, only you need to \textcolor{red}{get 4 in a row to win}. When playing, does putting my symbol in the middle guarantee me winning? No. With both players playing optimally, the result is always a draw. How many X's and O's do I need to play tic tac toe on a board game? Since the board itself has nine spaces, I recommend that you have nine for both X's and O's... \makecell[l]{ \url{https://www.wikihow.com/}\\\url{Play-Tic-Tac-Toe}}
            &
            ...~1. The game requires two players, X and O. 2. The game board is a set 3x3 grid in which players will place their symbol to claim that segment. 3. X typically players first, then players alternate turns. 4. The goal is to claim \textbf{three} \textcolor{red}{segments of the grid in a row}, either horizontally, vertically, or diagonally. 5. No additional sides can be added to the grid. 6. The game is over either when one player achieves \textcolor{red}{three segments in a row}, or when the grid is filled without anyone achieving three segments in a row.... \makecell[l]{ \url{https://www.siammandalay.com}\\\url{/blogs/puzzles/how-to-win-}\\\url{tic-tac-toe-tricks-to-always}\\\url{-win-noughts-crosses}} \\
            \bottomrule
	\end{tabular}
	\caption{Three qualitative example from TriviaQA, CommonsenseQA, and NumerSense. We present the top-1 retrieved passages from Wikipedia, CCNet, and web. The words in red denote the keywords related to the question.}
	\label{tab:app-case-study}
\end{table*}

\end{document}